\documentclass[twocolumn,3p,times]{elsarticle}
\usepackage[utf8]{inputenc}
\usepackage[T1]{fontenc}

\usepackage{natbib}\let\cite\citep
\usepackage{latexsym}
\usepackage{soul,color}

\usepackage{graphicx}
\usepackage{multirow}
\usepackage{amsmath}

\usepackage{epsdice} 
\usepackage{algorithm,algpseudocode} 
\usepackage{cleveref}

\newcommand{\dataset}{DepressionEmo}

\title{\dataset{}: A novel dataset for multilabel classification of depression emotions}

\usepackage{xurl} 

\usepackage{pbox}

\usepackage{amssymb}
\usepackage{svg}
\usepackage{booktabs}  
\usepackage{seqsplit}

\journal{Journal of Affective Disorder}

\begin{document}

\begin{frontmatter}






\author[aff1]{Abu~Bakar~Siddiqur~Rahman}
\ead{abubakarsiddiqurra@unomaha.edu}
\author[aff2]{Hoang-Thang Ta~}
\ead{tahoangthang@gmail.com}
\author[aff1]{Lotfollah~Najjar}
\ead{lnajjar@unomaha.edu}
\author[aff1]{Azad Azadmanesh}
\ead{azad@unomaha.edu}
\author[aff3]{Ali Saffet Gönül}
\ead{ali.saffet.gonul@ege.edu.tr}

\address[aff1]{College of Information Science and Technology, University of Nebraska Omaha, USA}
\address[aff2]{Department of Information Technology, Dalat University, Da Lat, Lam Dong, Vietnam}
\address[aff3]{Department of Psychiatry, Ege University, Bornova, Izmir, Turkey}

\begin{abstract}

Emotions are integral to human social interactions, with diverse responses elicited by various situational contexts. Particularly, the prevalence of negative emotional states has been correlated with negative outcomes for mental health, necessitating a comprehensive analysis of their occurrence and impact on individuals. In this paper, we introduce a novel dataset named DepressionEmo designed to detect 8 emotions associated with depression by 6037 examples of long Reddit user posts. This dataset was created through a majority vote over inputs by zero-shot classifications from pre-trained models and validating the quality by annotators and ChatGPT, exhibiting an acceptable level of inter-rater reliability between annotators. The correlation between emotions, their distribution over time, and linguistic analysis are conducted on DepressionEmo. Besides, we provide several text classification methods classified into two groups: machine learning methods such as SVM, XGBoost, and Light GBM; and deep learning methods such as BERT, GAN-BERT, and BART. The pretrained BART model, \texttt{bart-base} allows us to obtain the highest F1-Macro of 0.76, showing its outperform compared to other methods evaluated in our analysis. Across all emotions, the highest F1-Macro value is achieved by suicide intent, indicating a certain value of our dataset in identifying emotions in individuals with depression symptoms through text analysis. The curated dataset is publicly available at: \url{https://github.com/abuBakarSiddiqurRahman/DepressionEmo}.

\end{abstract}





\end{frontmatter}


\section{Introduction}\label{sec1}

Depression is a complex mental disorder that afflicts approximately 300 million people worldwide, which is about 1 in every 25 individuals experience the impact of depression~\cite{zhang2019evaluating}. It hinders patients' quality of life, represents a significant global health concern, and becomes a threat if one suffers from severe depression condition~\cite{morrill2008interaction}. Depression, stemming from emotions like emptiness, sadness, and anger, often leads to a communication barrier and increased suicide risk. This isolation can result in resentment towards others. The challenge lies in effective communication and seeking help. Social media, particularly platforms, has become a space for individuals to share their emotions. Recognizing varied emotions among depressed individuals is crucial, especially in the vast social media landscape. Natural Language Processing and deep learning are effective tools for automatically identifying these emotions, offering insights into factors contributing to suicidal thoughts among users.

In the field of text classification, depression detection in text often utilizes machine learning and deep learning approaches. Deep learning, in particular, tends to achieve superior performance overall due to its ability to capture the semantic relationships between words effectively, especially when training models with billions of parameters. Various methods for automated depression detection in text include word embedding and bi-LSTM with attention~\cite{Ren2021Depression}, Prompt-based Topic-modeling (PTDD)\cite{guo2023prompt}, multi-task learning\cite{wang2023cognitive}, LIWC with culture and suicide lexicons~\cite{lyu2023detecting}, knowledge-aware deep learning~\cite{zhang2023depression}, a deep learning framework with expectation loss~\cite{de2019depression}, Hierarchical Attention Network~\cite{han2022hierarchical}, and BERT, RoBERTA, and XLNET models~\cite{wang2020depression}. Recent studies have identified the utilization of large language models such as ChatGPT to work on different aspects of depression~\cite{heston2023safety,xu2023leveraging,lamichhane2023evaluation}. These diverse deep-learning methods advance automated depression detection in text.


Twitter data, as analyzed by~\citet{chiong2021textual}, has been employed for depression detection using machine learning classifiers with textual features. Similarly, depression detection studies using Facebook data have utilized user posts as predictive tools~\cite{al2019depression, katchapakirin2018facebook}. In contrast, some have focused on diagnosing depressive disorders from content shared on Facebook timelines~\cite{eichstaedt2018facebook}. Reddit, another popular platform, has been used to detect depression by analyzing textual contributions from individuals and control subjects~\cite{chen2023detecting}. Additionally, blog content, especially from platforms like Blogspot, contributes to identifying depressive tendencies through text analysis~\cite{nguyen2014affective}.

This paper introduces a novel dataset called \dataset{}, sourced from Reddit, aiming to explore emotions associated with depression in texts from specific subreddit posts. Unlike many existing datasets, \dataset{} uniquely identifies eight depression emotions within long Reddit posts. Annotation of \dataset{} was carried out using zero-shot classification on pre-trained models, relying on majority vote results. The evaluation involved both ChatGPT and human annotators to assess label quality, focusing on concordance through reliability metrics between human annotators and between annotators and ChatGPT against data labels. Some notable experiments included data analyses on depression distributions concerning raw frequencies, time, and emotion pair correlations. For depression detection, we applied various text classification methods, such as SVM, XGBoost~\cite{chen2015xgboost}, Light GBM~\cite{ke2017lightgbm}, BERT~\cite{devlin2018bert}, GAN-BERT~\cite{croce2020gan}, BART~\cite{lewis2019bart}, with the first three considered as baseline machine learning methods and the latter as deep learning methods. The primary contributions of this study are outlined below.
\begin{enumerate}[1.]
    \item To present an innovative multilabel dataset focused on depression-related texts extracted from Reddit, meticulously annotated with corresponding emotions. Additionally, we conducted various data analyses to gain insights into the distribution of emotions within the dataset.
    \item To employ widely used text classification techniques to identify emotions in textual content and subsequently compare their output model performances.
\end{enumerate}

The rest of this document is structured as follows: Section 2 discusses previous research on detecting depression in texts. Sections 3 and 4 outline the process of creating the \dataset{} and conducting various data analyses. Our multilabel classification methods are detailed in Section 5, and we apply them in the experiments outlined in Section 6. Section 7 addresses an error analysis and our work's limitations, which we provide in Section 8. Finally, we make conclusions and outline future work in Section 9.

\section{Related Works}

Major depression significantly impacts the lives of millions of individuals in the USA and globally. In their lifetime, a minimum of one in five adults in the US will encounter depression, with around 6\% experiencing it in any given year~\cite{hasin2005epidemiology}. Understanding the shifts in depression rates over time in the US is crucial due to its significant impact on a large segment of the population and its profound personal and societal effects. This insight is vital for effectively directing funding and public health strategies to address the community's current needs. Moreover, there has been a growing disparity in depression prevalence among at-risk groups in the US between 2005 and 2015~\cite{weinberger2018trends}. This growing disparity is further evidenced by a study indicating a significant prevalence of depression among Latinos in the United States~\cite{menselson2008depression}. Additionally, the correlation between depression and religiosity/spirituality in college students has been investigated, uncovering notable insights into the mental health challenges faced by this demographic~\cite{berry2011depression}. Furthermore, research focusing on the treatment of adult depression in the United States has shed light on various therapeutic approaches, highlighting the evolving landscape of mental health care~\cite{olfson2016treatment}. It is imperative to discern the underlying causes and specific emotional triggers that predominantly contribute to the onset of depression. 



To accomplish the automatic detection of depression, researchers used machine learning, deep learning methods, and Natural Language Processing. Recently, work has been done to detect depression using word embedding, and later, the word embedding fed to bi-LSTM and attention layer to understand the contextual information~\cite{Ren2021Depression}. The Prompt-based Topic-modeling method for Depression Detection (PTDD) utilizes pre-trained language models to address the challenges of limited training data by reorganizing text data into predefined topics for each interviewee. This method employs a prompt-based framework for predicting the emotional tone of next-sentence prompts, facilitating depression detection through topicwise predictions and a simple voting process~\cite{guo2023prompt}. A multi-task learning model based on a pre-trained model was used to detect depression~\cite{wang2023cognitive}.  Linguistic Inquiry Word Count (LIWC) was combined based on culture and suicide lexicons to detect depression from textual data~\cite{lyu2023detecting}. Knowledge-aware deep learning methods were used to detect depression~\cite{zhang2023depression}. A deep learning framework is designed to predict depression levels using distribution learning precisely. This framework is based on an innovative expectation loss function, enabling the estimation of data distribution across varying depression levels by optimizing expected values to align closely with actual ground-truth levels~\cite{de2019depression}. In some cases, the accuracy of existing models could have been better, primarily due to imbalanced data distributions in real-world scenarios. To address this issue, a deep learning model named X-A-BiLSTM was proposed for depression detection in imbalanced social media data from Reddit. This model integrates two key components: XGBoost, utilized to mitigate data imbalance, and an Attention-BiLSTM neural network, which boosts classification capabilities~\cite{cong2018xa}. Multi-Aspect Depression Detection with Hierarchical Attention Network was used to detect depression from social media~\cite{zogan2022explainable} automatically. BERT, RoBERTA, BERTweet, and mentalBERT were used to detect depression from Twitter and Reddit data~\cite{tavchioski2023detection}.


Depression detection work has already been done with specific goals, such as severity detection, early symptom detection, and summarization. Depression severities were detected from Twitter data~\cite{kabir2023deptweet}, and some work has been done using multimodal~\cite{stepanov2018depression}. Early detection was used by using Reddit data~\cite{trotzek2018utilizing, martinez2020early}. Late-life depression was detected using Natural Language Processing techniques~\cite{desouza2021natural}. Social media texts are a mixture of relevant and irrelevant content due to the goal of having only depression-related texts for detecting depression. Hybrid abstractive and extractive summarization are used by~\cite{zogan2021depressionnet} to collect relevant texts from Twitter to detect depression. BERT with extractive summarization was used to process data from Reddit and Twitter for detecting depression~\cite{william2022leveraging}.   

In most cases, depressed people suffer from sharing their emotions with others. Social media works as a perfect platform where they can share their expressions. Research conducted by various authors has utilized social media data to understand this phenomenon better. For example, a study extracted datasets from Twitter, including 292,564 tweets from 1,402 users identified as depressed, over 10 billion tweets from a broader base of more than 300 million users, and 35,076,677 tweets from 36,993 users categorized as potential depression candidates. This extensive data collection offers valuable insights into the social media behaviors of individuals with depression~\cite{shen2017depression}. The CLPsych shared task in 2015 provided a Twitter dataset for identifying users with PTSD, depression, and control users. This dataset was structured to include up to 3,000 tweets from each user while ensuring that a minimum of 25 tweets were available for each individual~\cite{coppersmith2015clpsych}. Building upon these research efforts, another study amalgamated the previously mentioned datasets with an additional Twitter dataset representative of Canada's population, further enhancing the scope and accuracy of depression detection~\cite{skaik2020using}. Complementing this, another research effort collected a total of 89,192 tweets, focusing on a random subset of 1,500 tweets to aid in detecting depression~\cite{owen2020towards}. Similarly, a study utilizing Reddit social media analyzed 400 subreddit posts for training and another 400 for testing, specifically to identify depression~\cite{pirina2018identifying}. Adding to this research, a comprehensive analysis involving 716,068 posts from 3,940 users identified with depression and an equal number of posts from control users, totaling 1,177,764, was conducted to enhance the understanding and detection of depression~\cite{chen2023detecting}. Reddit data is also used for early risk prediction of depressed users~\cite{losada2016test,losada2017clef}.  


\section{Dataset Creation}

\subsection{Data Crawler}
To create \dataset{}, we developed a web crawler to collect user posts from different subreddits (\texttt{r$\backslash$depression}, \texttt{r$\backslash$Depressed partners}, \texttt{r$\backslash$loneliness}, \texttt{r$\backslash$suicide}, and \texttt{r$\backslash$suicide\_watch}) from searching for various depression-related keywords. 

We considered the texts from subreddit \texttt{r$\backslash$depression} and \texttt{r$\backslash$Depressed partners} as depressed related texts. We found out some specific keywords and common words from subreddit \texttt{r$\backslash$depression} and \texttt{r$\backslash$Depressed partners} to gathered examples from \texttt{r$\backslash$loneliness}, \texttt{r$\backslash$suicide}, and \texttt{r$\backslash$suicide\_watch} to include in our dataset. We exploited various common pronouns like \texttt{I}, \texttt{my}, \texttt{me}, \texttt{im}, \texttt{i'm}, and \texttt{am}, absolute terms such as \texttt{everything}, \texttt{nothing}, \texttt{always}, \texttt{anymore}, and \texttt{any}, as well as negative expressions like \texttt{no} and \texttt{never}. Additionally, more specific keywords, including \texttt{lonely}, \texttt{loneliness}, \texttt{alone}, \texttt{suicide}, \texttt{fuc*}, \texttt{shi*}, \texttt{feel}, \texttt{depression}, \texttt{can't sleep}, \texttt{no sleep}, \texttt{feel exhausted}, and \texttt{no motivation} were considered for collecting data from subreddits \texttt{r$\backslash$loneliness}, \texttt{r$\backslash$suicide}, and \texttt{r$\backslash$suicide\_watch}. 

From around \texttt{8k} extracted examples, we reduced them to \texttt{6k} examples by considering post length with only 256 tokens and eliminating the examples without emotions or providing only mental health suggestions for users. 

\subsection{Data Preprocessing}

Each user post is comprised of six fields: \texttt{id}, \texttt{title}, \texttt{post}, \texttt{text}, \texttt{upvotes}, \texttt{date}, and \texttt{emotions}. The \texttt{text} field is constructed from the concatenation of \texttt{title} and \texttt{post}, using the delimiter $\#\#\#$, then detect depression emotions in it by using the majority vote from zero-shot classifications models, presented in \Cref{annotation}. To guarantee the appropriate length of sequences for training models, any text surpassing 240 tokens or falling below five tokens is omitted. This ensures that the models concentrate on training with a maximum of 256 tokens. In addition, we have done some basic preprocessing steps such as lemmatization and removing stop words and punctuation. We have no other preprocessing steps to keep the data original and to see how models deal with it. 

\subsection{Emotion Definitions}

While depression is associated with numerous emotions, this paper focuses on eight emotions that frequently emerge in text and exhibit the strongest correlation between annotators.

\noindent{\textbf{Anger}: Anger is an intensified emotional response that can be directed inward or outward. It contributes to feelings of despair and self-criticism that pivots in depression development, as individuals with depression grapple with navigating and expressing their anger~\cite{busch2009anger}.}

\noindent{\textbf{Cognitive dysfunction}: Cognitive dysfunction denotes an impairment in the cognitive's normal functioning, causing disruptions in cognitive, emotional, or behavioral processes. Forgetfullness, slow thinking, difficulties with expressing thoughts experienced with cognitive dysfunction ~\cite{jarema2014cognitive}.}

\noindent{\textbf{Emptiness}: Emptiness signifies a deep emotional void, numbness, or a lack of vitality, extending beyond typical sadness. Emptiness with depression experienced disconnected from others, and an empty future~\cite{d2020feeling}.}

\noindent{\textbf{Hopelessness}: The connection between hopelessness and depression is intrinsic; hopelessness serves as a susceptibility factor in the onset of depressive symptoms~\cite{abramson1989hopelessness}.}

\noindent{\textbf{Loneliness}: Loneliness is a heightened emotional state marked by profound isolation, even in the presence of others. Viewing loneliness as an adverse emotion was regarded as a predisposing element in the development of depression ~\cite{erzen2018effect}.}

\noindent{\textbf{Sadness}: Sadness is a natural human emotion often triggered by specific events or losses, both physically and mentally manifesting. Many authors recognize sadness as a fundamental symptom of depression~\cite{mouchet2008sadness}.}

\noindent{\textbf{Suicide intent}: Suicidal intent is the conscious desire to end one's life, indicating a serious mental health concern that demands immediate attention. Those with suicidal intent may be overwhelmed by emotional pain, and a belief that ending their life is the only escape from suffering. Suicidal ideation is also related to depression ~\cite{vangastel1997prediction}.}

\noindent{\textbf{Worthlessness}: Worthlessness is a profound sense of having little or no value, commonly associated with mental health conditions like depression ~\cite{zahn2015selfblame}. Those experiencing this emotion often view themselves as inadequate or undeserving of positive regard and going through an existential crisis where they might feel as if they don’t exist.}

\subsection{Annotation Process}\label{annotation}

Because human annotation comes with a high cost, we categorize eight emotions (anger, cognitive dys\-func\-tion, emptiness, hopelessness, loneliness, sadness, suicide intent, and worthlessness) within the \texttt{text} fields by leveraging zero-shot classification outcomes from pre-trained models. Addressing this as a multilabel classification challenge, we determine the final emotions through a majority vote. We use 4 pre-trained models (\texttt{model1 = \seqsplit{MoritzLaurer/mDeBERTa-v3-base-mnli-xnli}}, \texttt{model2 = \seqsplit{MoritzLaurer / DeBERTa-v3-large-mnli-fever-anli-ling-wanli}}, \texttt{model3 = \seqsplit{cross-encoder / nli-deberta-v3-small}}, and \texttt{model4 = \seqsplit{facebook / bart-large-mnli}}) to classify each \texttt{text} and annotate any emotion that appears at least 2 over the total of 4 model outputs (each model gives an output). For each pre-trained model, labels with scores greater than 0.5 are considered, and the corresponding emotions are added as the output from each model. Several chosen examples are presented in \Cref{tab:dataset_examples}.

\begin{table*}[!htb]
\small
\centering
\caption{Several chosen examples in \dataset{}.}

\begin{tabular}{p{7cm}lp{1.5cm}}
\hline
\textbf{Text (Title + Post} & \textbf{Date} & \textbf{Emotions}  \\
\hline
\texttt{it’s okay to cry $\#\#\#$ After 8 months of break up and 4 months of no contact, I finally cried for the first time. It was a strange kind of cry.... a cry that I knew I had no emotions and I do not want her back in my life (even after she dumped me through an email). To all that are going through a break up... you will have relapses and will go through different stages of break up... stay strong, be kind to yourself, speak to love ones and remember guys... it’s okay to cry.} &  2020-06-29 01:57:02 & sadness, emptiness, hopelessness \\
\hline
\texttt{How can I make tasks feel less big $\#\#\#$ I want to do three things today, I want to brush my teeth and clean my rabbits cage cage and I want to do my dishes, I feel scared it feels like so much to do and I’m sorry if that sounds so bad but that’s how I feel. I want it to feel like I can do it but it feels so big it feels like so much and I’m crying because I just want to go lay down and sleep again.} &  2022-06-29 20:56:31 & sadness, hopelessness \\
\hline
\texttt{Will it ever end? $\#\#\#$ The day to day loathing. The ideation of it being over. The sleeping through the day. The indecisiveness. The fear of being in public. The shame carried with depression. The lying. The pretending it’s okay. The exhaustion. I’d like to carry on but it won’t help. I can’t hold a job down, I’ve started carrying debt just so I don’t have to work. But when I run out of money, then I can count myself out. It will have to come to an end. I am ashamed of how pathetic I am.} &  2023-08-01 01:08:30 & hopelessness, sadness, worthlessness, suicide intent \\
\hline
\texttt{Does anyone else get the feeling that life is just passing them by? $\#\#\#$ I feel like time is moving too fast for me to seize it. I have no idea where 2018 went. And it was the same for 2017 and its gonna be similar for this year I fear.} &  2019-01-05 10:19:56 & sadness, hopelessness \\
\hline
\end{tabular}
\label{tab:dataset_examples}
\end{table*}

The annotation quality must be evaluated based on the inter-rater reliability between annotators. We follow a similar approach~\cite{ta2023wikides} but for checking the annotation quality. We asked 3 Ph.D. students to annotate 100 random examples and then measure their consensus with our dataset labeling by several coefficients, Bennett, Alpert and Goldstein's S (\textit{S})~\cite{bennett1954communications}, Fleiss' kappa or multi-kappa ($\kappa_{f}$)~\cite{davies1982measuring}, Krippendorff’s alpha ($\alpha$)~\cite{klaus1980content}, and Scott's Pi ($\pi$)~\cite{scott1955reliability}. These are frequently employed statistical measures for assessing the consistency among multiple evaluators when evaluating items across various categories. Below are the scales associated with each coefficient:
\begin{itemize}
    \item $\alpha$: The scale spans from -$1$ to $1$, where $1$ indicates perfect agreement, $0$ indicates no agreement beyond chance, and negative values indicate disagreement~\cite{zapf2016measuring}. $\alpha_{n}$ refers to alpha nominal that is used for nominal data. 
    \item $\pi$: The calculation is expressed as $\pi = P_o - P_e /(1-P_e)$, where $P_o$ represents the observed proportion of agreement and $P_e$ denotes the expected proportion by chance. The scale ranges from a minimum value of $-P_e/(1-P_e)$ (when $P_o$ equals $0$) to a maximum value of $1$ (when $P_o$ equals $1$)~\cite{craig1981generalization}.
    \item \textit{S}: The scale ranges from $-1$ to $1$. A perfect agreement between the two raters is denoted by $1$, while $-1/(n-1)$ is assigned when the observed agreement proportion $P_o$ equals $0$. The minimum value of $-1$ is associated with the number of categories $n$ being equal to $2$, and this value approaches $0$ as $n$ increases~\cite{warrens2012effect}.
    \item $\kappa_{f}$: The coefficient's author proposed that values $\leq$ 0 signify no agreement, 0.01–0.20 signify none to slight agreement, 0.21–0.40 signify fair agreement, 0.41–0.60 signify moderate agreement, 0.61–0.80 signify substantial agreement, and 0.81–1.00 signify almost perfect agreement. It is used for the agreement between 3 or more raters~\cite{mchugh2012interrater}.

\end{itemize}
\begin{table*}[!htb]
\small
\centering
\caption{The inter-rater reliability between annotators and dataset labels.}

\begin{tabular}{llllllll}
\hline
\textbf{Emotion} & $\alpha_{n}$  & $\pi$ & S & $\kappa_{f}$ \\
\hline
anger & 0.65  & 0.65 & 0.68 &  0.65 
\\
\hline
cognitive dysfunction & 0.94 & 0.93 & 0.97 & 0.94
\\
\hline
emptiness &  0.51 & 0.54 & 0.51 & 0.51 
\\
\hline
hopelessness &  0.64 & 0.64 & 0.67 & 0.64 
\\
\hline
loneliness & 0.62 & 0.64 & 0.62 & 0.61
\\
\hline
sadness & 0.97  & 0.97 & 0.98 & 0.97
\\
\hline
suicide intent & 0.92 & 0.91 & 0.94 & 0.92
\\
\hline
 worthlessness & 0.92 & 0.91 & 0.91 & 0.91
\\
\hline
\end{tabular}
\label{tab:inter_rater_reliability}
\end{table*}

The information presented in \Cref{tab:inter_rater_reliability} illustrates a fair agreement between annotators and dataset labels, with most coefficient values falling between 0.5 and 0.99. We consider these outcomes acceptable, considering the high cost of manual annotation and the challenges associated with multilabel classification. Additionally, in this context, the average value for sadness is the highest, while emptiness has the lowest average value.

\begin{table*}[!htb]
\small
\centering
\caption{The inter-rater reliability between annotators, ChatGPT bots, and dataset labels.}

\begin{tabular}{llllllll}
\hline
\textbf{Emotion} & $\alpha_{n}$ & $\pi$ & S & $\kappa_{f}$\\
\hline
anger &  0.73 & 0.95 & 0.95 & 0.73
\\
\hline
cognitive dysfunction &  0.64 & 0.94 & 0.94 & 0.64
\\
\hline
emptiness &  0.62 & 0.97 & 0.97 & 0.62
\\
\hline
hopelessness &   0.52 & 0.95 & 0.95 & 0.52
\\
\hline
loneliness &  0.53 & 0.93 & 0.52 & 0.53
\\
\hline
sadness &  0.61 & 0.93 & 0.63 & 0.61
\\
\hline
 suicide intent &  0.85 & 0.98 & 0.97 & 0.85
\\
\hline
 worthlessness &   0.56 & 0.97 & 0.97 & 0.56
\\
\hline
\end{tabular}
\label{tab:inter_rater_reliability_chatgpt}
\end{table*}


In \Cref{tab:inter_rater_reliability_chatgpt}, each emotion label extract for each annotator, dataset labels, and from ChatGPT to make it binary whether any specific emotion is selected by any of the above-mentioned annotations or not. The inter-rate agreement shows high agreement for all emotions, among them, anger, suicide intent, cognitive dysfunction and emptiness shows high agreement. 

The previous experiment centers on a subset of \dataset{} comprising 100 random examples; it is unclear whether the inter-rater reliability is the same for the whole dataset. Therefore, another experiment is necessary to prove the annotation quality, and we have opted for the odds ratio, which can evaluate the agreement among annotators.

\begin{table*}[htb]
\small
\centering
\caption{Contingency tables and odds ratios between annotators by emotions.}
\begin{tabular}{lcccccc}
\hline
\textbf{100 random examples} & \textbf{A. 1} & \textbf{A. 2} & \textbf{A. 3} & \textbf{OR (1/2)} & \textbf{OR (1/3)} & \textbf{OR (2/3)}\\
\hline
with anger & 33 & 31 & 37 & & & \\
without anger & 67 & 69 & 63 & 1.09 & 0.83 & 0.76 \\
\hline
with cognitive dysfunction & 16 & 13 & 14 & & & \\
without cognitive dysfunction & 84 & 87 & 86 & 1.27 & 1.17 & 0.917 \\
\hline
with emptiness & 49 & 45 & 42 & & & \\
without emptiness & 51 & 55 & 58 & 1.17 & 1.32 & 1.12 \\
\hline
with hopelessness & 65 & 59 & 67 & & & \\
without hopelessness & 35 & 41 & 33 & 1.29 & 0.91 & 0.70\\
\hline
with loneliness & 67 & 45 & 44 & & & \\
without loneliness & 33 & 55 & 56 & 2.48 & 2.58 & 1.04 \\
\hline
with sadness & 79 & 77 & 78 & & & \\
without sadness & 21 & 23 & 22 & 1.12 & 1.06 & 0.95 \\
\hline
with suicide intent & 25 & 21 & 24 & & & \\
without suicide intent & 75 & 79 & 76 & 1.25 & 1.05 & 0.84 \\
\hline
with worthlessness & 51 & 48 & 55 & & & \\
without worthlessness & 49 & 52 & 45 & 1.12 & 0.85 & 0.75 \\
\hline
with all emotions & 385 & 320 & 361 & & & \\
without all emotions & 415 & 480 & 439 &1.39 & 1.12 & 0.82 \\
\hline
\multicolumn{7}{l}{A. 1, A. 2, and A. 3 refer to Annotator 1, Annotator 2, and Annotator 3, respectively.} \\
\multicolumn{7}{l}{OR (1/2), OR (1/3), and OR (2/3) refer to the Odds Ratios for Annotator 1 vs. 2, 1 vs. 3,} \\
\multicolumn{7}{l}{and 2 vs. 3, respectively.} \\
\hline
\end{tabular}
\label{tab:consistancy_annotators}
\end{table*}

We merge annotations from multiple annotators to form a matrix, with each row representing an example and each column denoting a distinct emotion. Each annotator maintains their own matrix. Following this, a contingency table is established by counting occurrences of examples associated with a specific emotion (e.g., anger) and those without it. Let's define an annotator $A$, who annotates an emotion $e$ by determining the count of examples with $e$ ($x$) and those without $e$ ($y$). The odds ratio $Or$ between two annotators $A_1$ and $A_2$ over an emotion $e$ is defined by:

\begin{equation}
\begin{aligned}
Or(A_1, A_2, e) &= \frac{\text{Odds of occurrences of examples with } e}{\text{Odds of occurrences of examples without } e} \\
&= \frac{x_1/x_2}{y_1/y_2} = \frac{x_1 \times y_2}{y_1 \times x_2}
\end{aligned}
\label{eq:odds_ratio}
\end{equation}

The odds ratios and contingency tables for different emotions are shown in \Cref{tab:consistancy_annotators}. For example, anger has an odds ratio of 1.09 between annotator 1 vs. annotator 2, computed as \(\frac{33 \times 69}{31 \times 67}\) using \Cref{eq:odds_ratio}. All emotions demonstrate values exceeding 1, indicating consistency among annotators.

if, $OR = 1$: Indicates perfect agreement between annotators. The odds of one annotator assigning a particular rating are the same as the odds of the other annotator assigning that rating.
$OR > 1$: Indicates that the first annotator is more likely to assign the rating in 100 samples than the second annotator.
$OR < 1$: Indicates that the first annotator is less likely to assign the rating in 100 samples than the second annotator.

Overall, there is consistency between annotators for all emotions except for the loneliness emotion annotator 1 with annotator 2 and annotator 3. However, the annotation is followed consistency based on the total number of emotions.

\subsection{Dataset Split}
\begin{figure}[!htb]
  \centering
  \includegraphics[width=\linewidth]{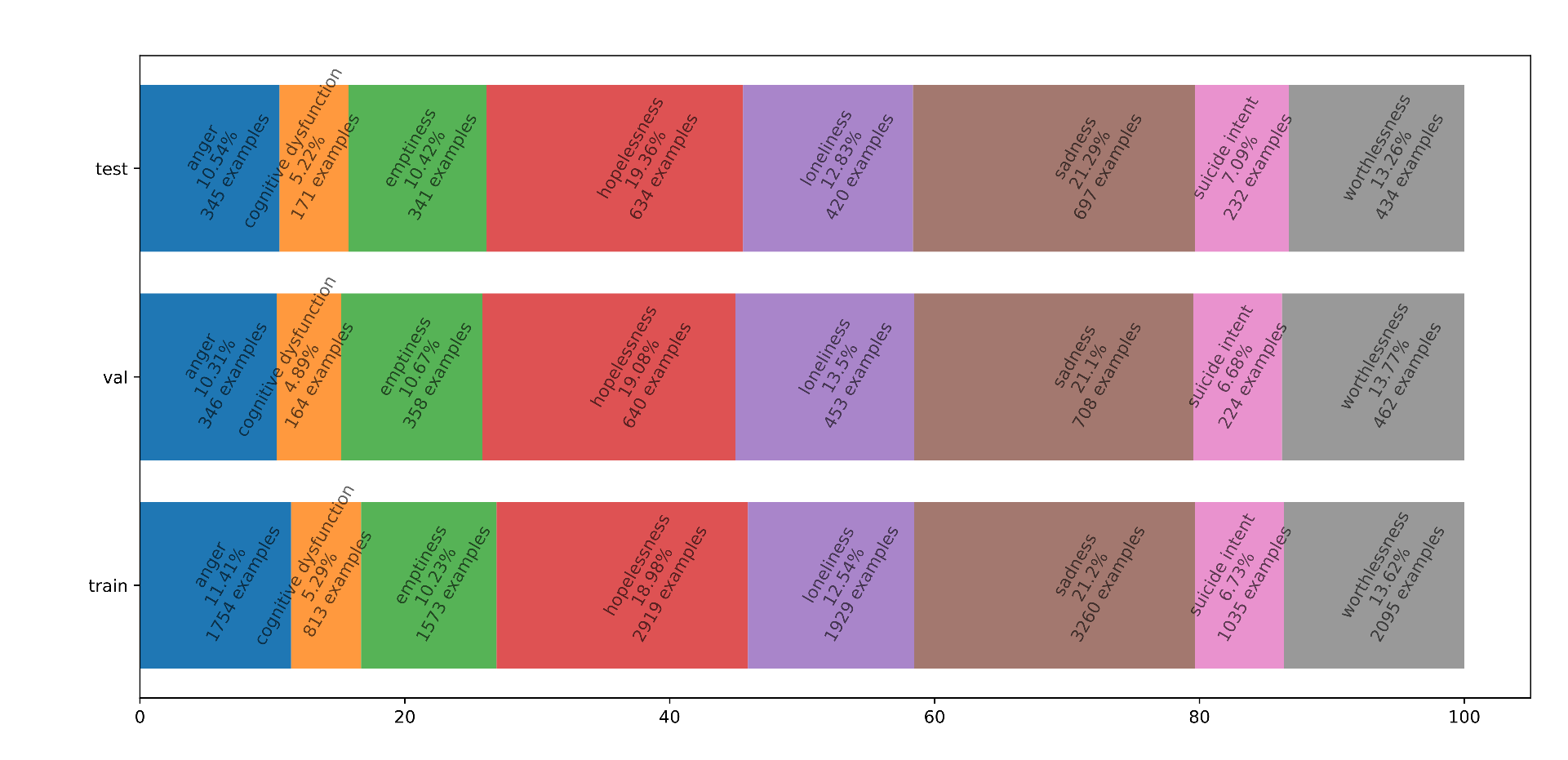}
  \caption{Distribution of emotions, by percentage and number of examples, across the training, validation, and validation sets}
  \label{fig:emotion_subset_distribution}
\end{figure}

The dataset is randomly divided into training (4225 examples), validation (906 examples), and test sets (906 examples) with a ratio of 7:1.5:1.5. This division is implemented to guarantee an equal distribution of all emotions across training, validation, and test sets. \Cref{fig:emotion_subset_distribution} shows the distributions by each emotion, including the percentage and the number of examples in the training, validation, and test sets. Overall, the distribution of examples related to emotions within datasets is relatively uniform, guaranteeing that the model is exposed to ample examples of each emotion throughout the training process.

\section{Dataset Exploration}

\subsection{Basic Analysis}

In this section, we conduct various basic analyses on \dataset{} to comprehend text length distribution and emotions' distributions by time.
After concatenating \texttt{title} and \texttt{post} as \texttt{text} for each user, we see that the text length distribution from 5 to 240 is roughly evenly distributed. This distribution is helpful enough for training models on a maximum input length of 256 tokens.

\begin{figure}[htb]
  \centering
\includegraphics[width=\linewidth]{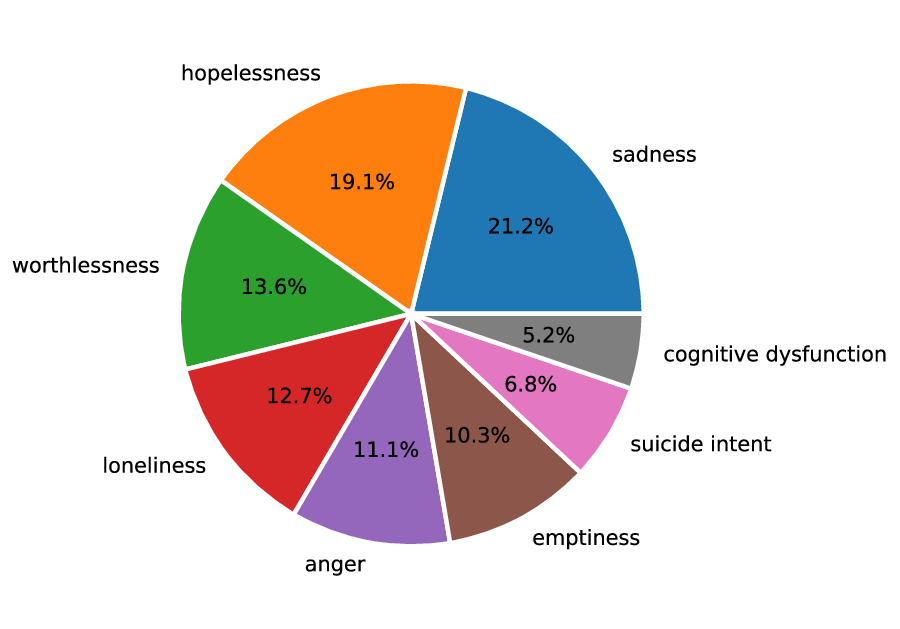}
  \centering
  \caption{Emotion distribution.}
\label{fig:emotion_distribution}
\end{figure}

\begin{figure}[!htb]
  \centering
\includegraphics[width=\linewidth]{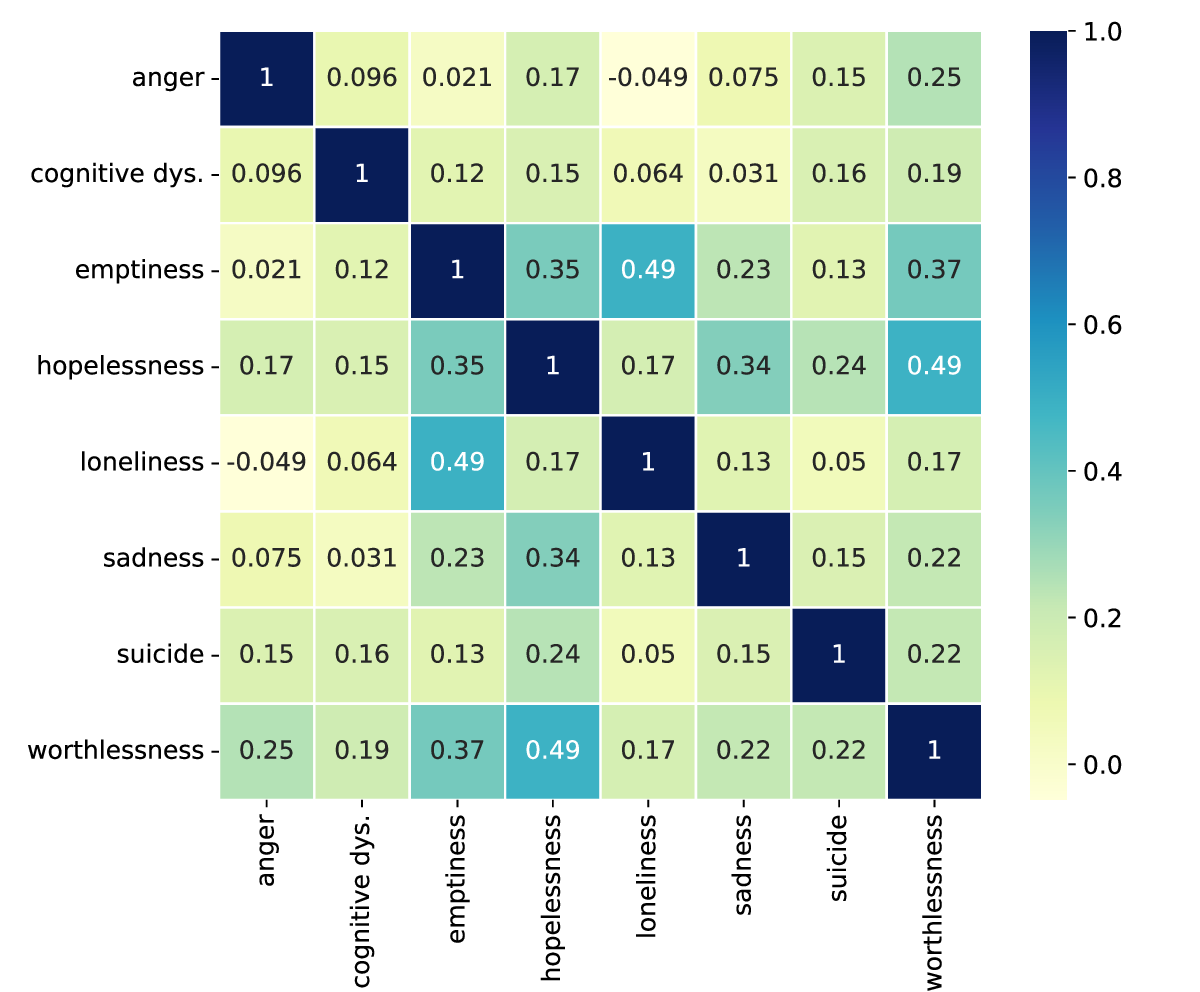}
  \centering
  \caption{The heatmap shows the Pearson correlation of emotion pairs.}
\label{fig:emotion_correlation}
\end{figure}

\Cref{fig:emotion_distribution} illustrates the emotional distribution across texts, with sadness comprising the largest share at 21.2\%, followed by hopelessness, while cognitive dysfunction represents the smallest portion at 5.2\%. Analyzing this distribution, we computed a Shannon entropy of 0.52, signifying that \dataset{} exhibits some imbalance, although within an acceptable range. The Shannon entropy scale ranges from 0 to 1, where 0 denotes an unbalanced dataset and 1 indicates a balanced one. 

\Cref{fig:emotion_correlation} shows the correlation between the pair of emotions. It indicates that the correlation is high between hopelessness and worthlessness. Hence, it is likely to appear together for a sample in a multilabel classification. On the contrary, anger and loneliness exhibit minimal correlation, suggesting that they are less likely to co-occur in a text. The heatmap also provides insight into the mutual exclusivity of emotions in the dataset. \Cref{tab:keyword_by_emotion} presents the distribution of examples by the number of emotions in the training set. While there is a noticeable decrease in the number of examples as the count of emotions increases from 1 to 5, this decline becomes more pronounced for examples with 6 to 8 emotions, with the fewest examples featuring all 8 emotions.  

\begin{figure}[!htb]
  \centering
\includegraphics[width=\linewidth]{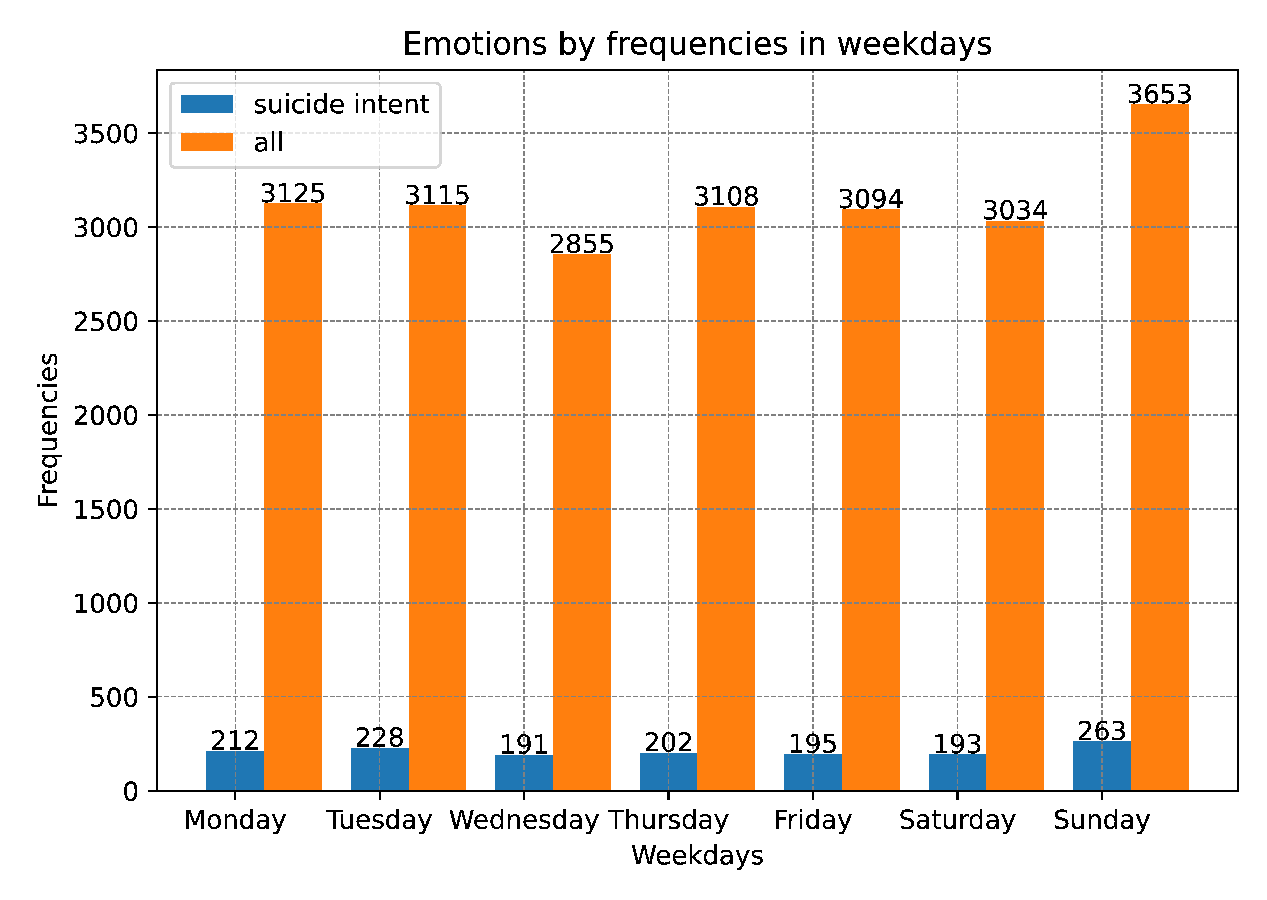}
  \centering
  \caption{Total emotions and ``suicide intent" emotions in weekdays.}
\label{fig:emotion_by_weekday_combined}
\end{figure}

\begin{figure}[!htb]
  \centering
\includegraphics[width=\linewidth]{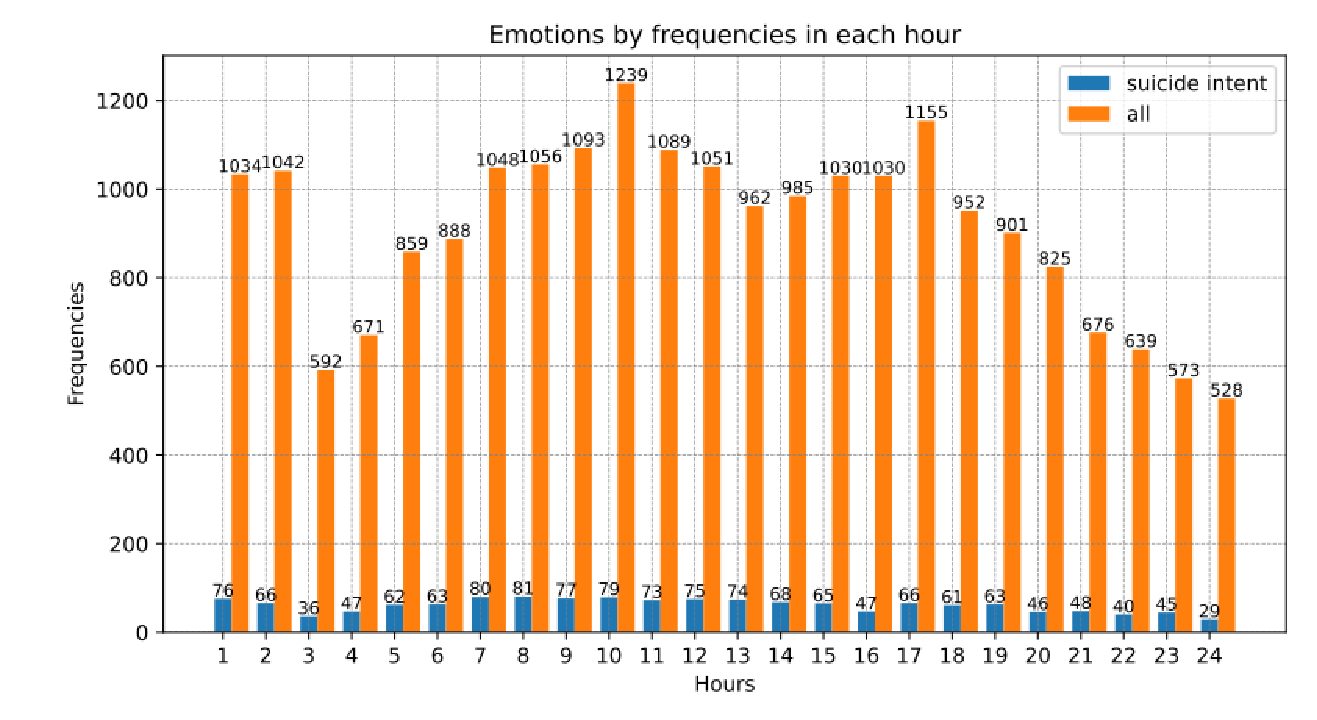}
  \centering
  \caption{Total emotions and ``suicide intent" emotions in 24 hours.}
\label{fig:emotion_by_hour_combined}
\end{figure}

The distributions of total emotions and ``suicide intent'' emotions across weekdays are displayed in   \Cref{fig:emotion_by_weekday_combined}. Sunday stands out as the day with the highest emotional expressions, whereas Wednesday emerges as the ``happiest'' day of the week. This is also true for ``suicide intent'' emotions. \Cref{fig:emotion_by_hour_combined} illustrates the distributions of total emotions and ``suicide intent'' emotions across 24 hours. In this analysis, we count the time from 1:00:00 AM to 1:59:59 AM as 1, similar to 2 for time from 2:00:00 AM to 2:59:59 AM, and so on. Notably, the lowest number of emotions occurs at 24 (from 24:00:00 to 24:59:59), closely followed by 3 (from 3:00:00 AM to 3:59:59 AM), while the peak is observed at 10 (from 10:00:00 AM to 10:59:59 AM). Similarly, the lowest number of ``suicide intents'' is also at 3, followed by 24.

\begin{table*}[!htb]
\small
\centering
\caption{Number of examples by each emotion in the training set.}
\begin{tabular}{p{5cm}p{4cm}}
\hline
\textbf{Emotions} & \textbf{Number of examples} \\
\hline
examples with 1 emotion & 785\\
examples with 2 emotions & 659\\
examples with 3 emotions & 659\\
examples with 4 emotions & 656\\
examples with 5 emotions & 633\\
examples with 6 emotions & 443\\
examples with 7 emotions & 269\\
examples with 8 emotions & 121\\
\hline
\end{tabular}
\label{tab:keyword_by_emotion}
\end{table*}

\begin{table*}[!htb]
\small
\centering
\caption{Statistical significance shown between LIWC features and emotions.}
\begin{tabular}{p{3cm}p{4cm}p{5cm} }
\hline
\textbf{LIWC categories} & \textbf{Words} & \textbf{Emotions} \\
\hline
1st person singular & i, me, my, myself & Anger***, Cognitive Dysfunction (forget)***, Emptiness***, Hopelessness***, Loneliness***, Sadness***, Suicide Intent***, Worthlessness***\\
All or none & all, no, never, always & Anger***, Emptiness***, Hopelessness***, Sadness***, Suicide Intent**, Worthlessness***\\
Negative tone & bad, wrong, too much, hate & Anger***, Cognitive Dysfunction (forget)***, Hopelessness***, Sadness***, Suicide Intent**, Worthlessness***\\
Negative emotions & bad, hate, hurt, tired & Anger***, Cognitive Dysfunction (forget)***, Emptiness***, Hopelessness***, Sadness***, Suicide Intent***, Worthlessness***\\
Swear & shit, fuckin*, fuck, damn & Anger***, Cognitive Dysfunction (forget)**, Suicide Intent***, Worthlessness***\\
Social behavior & said, love, say, care & Anger***, Sadness**\\
Illness & hospital*, cancer*, sick, pain & Hopelessness***, Sadness**, Suicide Intent***, Worthlessness***\\
Death & death*, dead, die, kill & Anger***, Cognitive Dysfunction (forget)***, Hopelessness***, Sadness***, Suicide Intent***, Worthlessness***\\
Fatigue & tired, bored, don’t care, boring & Anger***, Cognitive Dysfunction (forget)***, Emptiness***, Hopelessness***, Loneliness***, Sadness***, Suicide Intent***, Worthlessness***\\
Feeling & feel, hard, cool, felt & Cognitive Dysfunction (forget)**, Emptiness***, Hopelessness***, Loneliness***, Sadness***, Worthlessness***\\
\hline
\multicolumn{3}{l}{\textbf{***}: P value $<$ 0.001, \textbf{**}: P value $<$ 0.5} \\
\multicolumn{3}{l}{\textbf{*}: Wildcard Character} \\
\hline
\end{tabular}
\label{tab:p value}
\end{table*} 

The next analysis was considered using LIWC, an acronym for Linguistic Inquiry and Word Count, a specialized text analysis tool designed to categorize and quantify linguistic patterns\footnote{https://www.liwc.app/}. It aids in discerning the emotional, cognitive, and social dimensions of textual data. The primary purpose of using LIWC is to gain a deeper understanding of texts' psychological and emotional content. LIWC software categorizes words into specific linguistic and psychological dimensions, allowing researchers to quantify and analyze patterns in language use.

To assess the linguistic aspects based on \Cref{tab:p value}, a Pearson correlation analysis was conducted between LIWC features extracted from the combined text of titles and posts, and the dataset, considering each emotion label. The titles and posts were treated as text, incorporating linguistic dimensions, psychological processes, social processes, physical aspects, motives, and time-related categories based on 23 LIWC categories chosen for the analysis~\cite{boyd2022development}. The chosen 23 LIWC categories serve as a framework to analyze the linguistic content within the dataset systematically. LIWC features were employed to quantify the occurrence of words within the text associated with the dataset. Emotions were represented in a binary format, indicating whether they were linked to a particular example. Subsequently, p values were computed for each emotion label, providing insights into the statistical significance of the correlation with LIWC features based on word count in the dataset. We use two hypotheses:

\begin{itemize}
    \item $H_0$: LIWC features and emotion labels have no significant positive correlation.
    \item $H_a$: LIWC features and emotion labels correlate significantly positively.
\end{itemize}

A positive correlation between a specific LIWC category and an emotion suggests that a higher count of words associated with that LIWC category is associated with a higher likelihood of the presence of that emotion. In the context of the hypotheses, a significant positive correlation would indicate that there is a statistically meaningful relationship between the linguistic features captured by LIWC and the occurrence of specific emotions. The p values of each correlation coefficient quantify the evidence against the null hypothesis. A lower p value suggests stronger evidence against the idea that there is no positive correlation.

In \Cref{tab:p value}, p value $<$ 0.001*** and p value $<$ 0.05** were considered where each emotion showed by corresponding *** and **. For example, Anger*** with 1st person singular means the p value is less than 0.001 in that case. Hence, we can reject the null hypothesis, and it concludes that LIWC features and angry emotion labels have significant positive correlations and ensure statistical significance between them. We did this analysis with 23 LIWC categories. Among them, 10 are shown in the table. The others, such as the politeness LIWC category, show no correlation with any of the emotions, and some others such as the Need categories, show statistical significance with emptiness.  Furthermore, \Cref{tab:p value} shows:
\begin{itemize}
    \item The psycholinguistic analysis of the dataset shows which words are mostly aligned with depression texts and, in a broader sense, what type of emotions are mostly used by depressed users by using which categories of words. 
    \item Self-centered words, negative emotions, swear, death, these types of categories used mostly compared to politeness, love, or friendly communication categories. 
\end{itemize}

\subsection{Comparison with Other Depression Datasets}

To provide a rationale for establishing the \dataset{} for multilabel classification, we perform a comparative analysis with datasets that specifically focus on depression. This comparison assesses various features, including subset distribution (training/validation/test sets), source, text length, number of labels, and vocabulary size, as detailed in \Cref{tab:dataset_comparision}. For datasets with lengthy names, we assign our own abbreviated names. The computation of text length and vocabulary size for DATD, \dataset{}, and GoEmotions is based on spaCy\footnote{https://spacy.io/}, taking into account the removal of punctuation. We exclusively utilize word splitting from the clean text of other datasets, such as BLDC, MLDC, and SNCDL, in cases where using spaCy yields no difference.

\begin{table*}[!htb]
\small
\centering
\caption{The comparison of \dataset{} with other depression datasets.}
\resizebox{\textwidth}{!}{
\begin{tabular}{p{3cm}p{2.5cm}lp{1cm}p{1cm}p{1.2cm}}
\hline
\textbf{Dataset} & \textbf{Train/Val/Test} & \textbf{Source} &  \textbf{Text Len.} & \textbf{No. Labels} & \textbf{Vocab. Size}  \\
\hline
BLDC~\cite{nassar2022binary} & 60k & Twitter & 6.50 & 2 & 13,951 \\
DATD~\cite{owen2020towards} & 4650 & Twitter  & 11.24 & 2 & 9595
\\
\dataset{} (ours) & 4225/906/906 &  Reddit  & \textbf{103.97} & 8 & 18,192 \\
GoEmotions~\cite{demszky2020goemotions} & 58k & Reddit  & 12.99 & 27 & \textbf{68,937} \\

MLDC~\cite{radwa2022multi} & 57k & Twitter & 4.93 & 3 & 12,647 \\
SNCDL~\cite{haque2021deep} & 1850 & Reddit  & 82.98 & 2 & 11,869 \\
\hline
8datasets~\cite{pirina2018identifying} & 3600 & Forums and Reddit  & - & 2 & - \\
Depression~\cite{chen2023detecting}  & 1893K & Reddit  & - & 2 & - \\
\hline


\end{tabular}
}
\label{tab:dataset_comparision}
\end{table*}

All the datasets under consideration gather user posts from social media platforms like Twitter and Reddit. Notably, BLDC, DATD, MDLC, SNCDL, 8datasets, and Depression feature a limited number of labels (e.g., depression, no depression) and short text lengths. Most of the datasets collectively focus on binary classification. In contrast, GoEmotions stands out with a remarkable 27 different emotions, including non-depressive emotions like surprise and amusement, making it the most diverse dataset with a vocabulary size of 68k.


In a departure from other datasets, \dataset{} stands out by having the lengthiest average text at 103.97 and uniquely encompasses all emotions associated with depression. Like GoEmotions, \dataset{} is designed for multilabel classification but is characterized by fewer examples. To the best of our knowledge, \dataset{} may be among the few datasets dedicated to exploring depression-related emotions through multilabel classification.

\section{Methodology}

\subsection{Task Description}

As a multilabel classification problem, the task is to detect a list of emotions $E={e_1, e_2,...,e_N}$ occurring in a given text $t$, which can contain from 1 to N emotions, but it is not empty. In \dataset{}, there are 8 emotions, including anger, cognitive dysfunction, emptiness, hopelessness, loneliness, sadness, suicide intent, and worthlessness. 

\subsection{Methods}

Instead of training each model for each emotion in a setup of binary classification, we prefer to design a unified model that can detect all emotions at once in the inference time. There are two groups of methods: machine learning and deep learning.

For machine learning methods, we pick SVM~\cite{joachims1998text}, Light GMB~\cite{ke2017lightgbm}, and XGBoost~\cite{chen2015xgboost,chen2016xgboost}, which convert input data as numerical vectors by TF-IDF (Term Frequency-Inverse Document Frequency) for text classification. SVM depends on support vectors, which are key data points at the boundary of a given dataset, to determine the optimal hyperplane between classes. Additionally, it employs a kernel function to manage non-linear relationships~\cite{patle2013svm}. 

XGBoost considers sparsity, utilizing the Gradient Boosting Decision Tree (GBDT) principles for sparse data and weighted quantile sketch for approximating tree learning. By incorporating cache access patterns, data compression, and sharding, XGBoost efficiently handles billions of examples while using significantly fewer resources than comparable systems~\cite{chen2016xgboost}. Surpassing XGBoost in computational speed and memory efficiency while maintaining comparable model accuracy, LightGBM utilizes techniques such as Gradient-based One-Side Sampling (GOSS) and Exclusive Feature Bundling (EFB). GOSS removes data points with small gradients, achieving competitive accuracy with a reduced data size. Meanwhile, EFB groups mutually exclusive features to decrease the overall number of features~\cite{ke2017lightgbm}.



Although machine learning methods have proven beneficial for text classification, deep learning methods enhance model performance to a greater extent by employing millions of parameters during the training of neural networks. ~\citet{devlin2018bert} introduced BERT, a model that trains deep bidirectional representations by leveraging left and right context from unlabeled text. The pre-trained BERT model can demonstrate its efficacy in various natural language processing (NLP) tasks through fine-tuning with a single additional output layer. Nonetheless, BERT tends to excel primarily in medium and large datasets. In the case of \dataset{}, characterized as a small dataset with approximately 553 examples per emotion in the training set, we opt for GAN-BERT~\cite{croce2020gan}. GAN-BERT extends BERT by incorporating unlabeled data within a generative adversarial setting, aiming to showcase its effectiveness on small datasets. 

Ultimately, the absence of generative models for text classification would be a significant oversight. In this paper, we choose a generative model like BART, a denoising autoencoder utilized for pretraining sequence-to-sequence models~\cite{lewis2019bart}. BART undergoes training by introducing noise to the text through an arbitrary noising function, teaching the model to reconstruct the original text. Although BART is well-regarded for its success in text generation and comprehension tasks, we aim to assess its performance in multilabel classification by training it on the \dataset{}.

\section{Experiments}
We employ six methods to train our dataset, categorized into two groups: machine learning (ML) methods and deep learning (DL) methods. For SVM, Light GBM, and XGBoost, we utilize TF-IDF with default maximum features to vectorize inputs, subsequently using these vectors for model training. Furthermore, we use no text preprocessing steps because they obtain better output performance. Light GBM and XGBoost share a learning rate of 0.5. XGBoost is configured with parameters \texttt{max\_depth = 8} and \texttt{n\_estimators=100}, whereas SVM and Light GBM utilize default settings for the parameters.



Transfer learning is implemented for deep learning methods by utilizing small-sized pretrained models, specifically \texttt{bert-base-cased} for BERT and GAN BERT, and \texttt{bart-based} for BART. To make a fair comparison in deep learning models, GAN BERT, BERT, and BART use the same parameters, \texttt{training\_epoch = 25} and \texttt{max\_source\_length = 256}. The difference is GAN BART uses \texttt{batch\_size = 8} because it trains the Discriminator on Multi-Layer Perceptron, while BERT and BART use \texttt{batch\_size = 4}. In GAN BERT, we also distort noise data with normal distribution \texttt{N(-1, 1)} as mentioned in the work of ~\citet{ta2022gan}.


In \Cref{tab:test_set_results}, the test set results are presented across six metrics: F1 Macro, Precision Macro, Recall Macro, F1 Micro, Precision Micro, and Recall Micro. F1 Micro reflects the model's overall performance without consideration of label classes (emotions), while F1 Macro represents the average F1 values across all classes, with formulas: 

\begin{equation}
    \text{F1}_{\text{macro}} = \frac{1}{N} \sum_{i=1}^{N} \frac{2 \cdot \text{Precision}_{i} \cdot \text{Recall}_{i}}{\text{Precision}_{i} + \text{Recall}_{i}}
\label{eq:f1_macro}
\end{equation}
where, $N$ is the number of classes.

\begin{equation}
    \text{F1}_{\text{micro}} = \frac{2 \cdot \text{TP}_{\text{total}}}{2 \cdot \text{TP}_{\text{total}} + \text{FP}_{\text{total}} + \text{FN}_{\text{total}}}
\label{eq:f1_micro}
\end{equation}
where, $TP$ = True Positive, $FP$ = False Positive, and $FN$ = False Negative. 

Both F1 macro and F1 micro are suitable metrics for multi-class classification problems. F1 macro treats each class equally. It calculates the F1 score for each class independently and then averages them. This is particularly useful when there is a significant class imbalance, ensuring that each class contributes equally to the overall score. F1 micro aggregates the contributions from all classes into a single metric. It helps when there is a need to emphasize the performance of the majority class without neglecting the minority classes. 

The experimental findings indicate that DL methods outperform ML methods because they capture semantic meaning between words, as shown in \Cref{tab:test_set_results}. Generally, BART achieves the most favorable outcomes with an F1 Macro of 0.76 and an F1 Micro of 0.80. Notably, SVM exhibits superior precision values but lags behind in other metrics. From the experimental results, the DL methods outperformed ML methods because they can capture the semantic meaning between words in text. Generally, BART obtains the best results, with 0.76 for F1 Macro and 0.80 for F1 Micro. Surprisingly, SMV has the best precision values and the worst results in other metrics.

\begin{table*}[!htb]
\small
\centering
\caption{The results on the test set by different text classification methods.}
\resizebox{\textwidth}{!}{
\begin{tabular}{llllllllll}
\hline
\textbf{Method} & \textbf{Type} & \textbf{F1-Mac} & \textbf{P-Mac} & \textbf{R-Mac} & \textbf{F1-Mic} & \textbf{P-Mic} & \textbf{R-Mic} & \textbf{Avg.} \\
\hline
SVM & ML & 0.47 & \textbf{0.72}  & 0.41 &  0.61 & \textbf{0.77} & 0.51 & 0.58 \\
Light GBM & ML & 0.58 & 0.48  & 0.80 & 0.65 & 0.52 & 0.86  & 0.65 \\
XGBoost & ML & 0.59 & 0.63 & 0.56  & 0.66 & 0.69 & 0.63 & 0.63 \\
\hline
GAN-BERT & DL & 0.70 & 0.69  & 0.72 & 0.75 & 0.73 & 0.77 & 0.73 \\
BERT & DL & 0.74 & 0.72  & 0.77 & 0.79 & 0.76 & 0.83 & 0.77 \\
BART & DL & \textbf{0.76} & 0.70  & \textbf{0.81} & \textbf{0.80} & 0.74 & \textbf{0.86} & \textbf{0.78} \\ 
\hline
\multicolumn{8}{l}{\textbf{ML}: Machine Learning, \textbf{DL}: Deep Learning} \\
\multicolumn{8}{l}{\textbf{F1-Mac}: F1 Macro, \textbf{P-Mac}: Precision Macro, \textbf{Re-Mac}: Recall Macro } \\
\multicolumn{8}{l}{\textbf{F1-Mic}: F1 Micro, \textbf{P-Mic}: Precision Micro, \textbf{Re-Mic}: Recall Micro } \\
\multicolumn{7}{l}{\textbf{Avg.}: Average } \\
\hline
\end{tabular}
}
\label{tab:test_set_results}
\end{table*}

\Cref{tab:test_set_results_emotions} shows the results of each emotion over the test set on BART, the best method. We treat each emotion detection as the problem of binary classification. Therefore, there are two classes, 0 (no emotion) and 1 (emotion), with macro and micro metrics. On average, suicide intent gains the best metric values, and Cognitive dysfunction has the worst. This reflects the helpfulness of \dataset{} when showing the best performance in detecting suicide intent, a key emotion that leads to death in patients with depression.

\begin{table*}[!htb]
\small
\centering
\caption{The test set's result by emotions on BART.}

\resizebox{\textwidth}{!}{
\begin{tabular}{llllllll}
\hline
\textbf{Emotion} & \textbf{F1-Mac} & \textbf{P-Mac} & \textbf{R-Mac} & \textbf{F1-Mic} & \textbf{P-Mic} & \textbf{R-Mic}  & \textbf{Avg.}   \\
\hline
anger & 0.78 & 0.78 & 0.78 & 0.79 & 0.79 & 0.79 & 0.79 \\
\hline
cognitive dysfunction & 0.67 & 0.66 & 0.69 & 0.78 & 0.78 & 0.78 & 0.73 \\
\hline
emptiness & 0.76 & 0.76 & 0.78 & 0.77 & 0.77 & 0.77 & 0.77 \\
\hline
hopelessness & 0.73 & 0.81 & 0.70 & 0.80 & 0.80 & 0.80 & 0.77 \\
\hline
loneliness & 0.81 & 0.82 & 0.82 & 0.81 & 0.81 & 0.81 & 0.81 \\
\hline
sadness & 0.67 & 0.77 & 0.64 & 0.82 & 0.82 & 0.82 & 0.76 \\ 
\hline
suicide intent & \textbf{0.85} & \textbf{0.85} & \textbf{0.86} & \textbf{0.89} & \textbf{0.89} & \textbf{0.89} & \textbf{0.87} \\ 
\hline
worthlessness & 0.75 & 0.76 & 0.76 & 0.75 & 0.75 & 0.75 & 0.75 \\ 
\hline
\multicolumn{7}{l}{\textbf{F1-Mac}: F1 Macro, \textbf{P-Mac}: Precision Macro, \textbf{Re-Mac}: Recall Macro } \\
\multicolumn{7}{l}{\textbf{F1-Mic}: F1 Micro, \textbf{P-Mic}: Precision Micro, \textbf{Re-Mic}: Recall Micro } \\
\multicolumn{7}{l}{\textbf{Avg.}: Average } \\
\hline
\end{tabular}
}
\label{tab:test_set_results_emotions}
\end{table*}

\section{Error Analysis}

In this analysis, we conduct two methods: first, population proportion analysis to evaluate how closely the annotations from different sources align with the true labels of the dataset; second, confusion matrix analysis to assess \texttt{True Positives (TP)}, \texttt{True Negatives (TN)}, \texttt{False Positives (FP)}, and \texttt{False Negatives (FN)}, thereby comparing various annotations against the dataset's true labels for each emotion. This dual approach, encompassing both population proportion analysis and confusion matrix analysis, serves as a foundation for a comprehensive error analysis. Population proportion shows significant differences between the emotions of ``hopelessness'' and ``loneliness''. We conduct a confusion matrix analysis to understand the error more precisely for these two emotions. The confusion matrix analysis also provides some insightful conclusions that describe the part of the section below. 

 Population proportion analysis evaluates distinct hypotheses for each emotion, focusing on \texttt{10} pairwise comparisons across five labeling sources. Specifically, we calculated $\binom{5}{2}$ combinations, representing five unique labeling procedures: the true labels of our dataset (\texttt{t}), and annotations from three individual annotators (\texttt{a1}, \texttt{a2}, \texttt{a3}) along with ChatGPT (\texttt{g}). The \texttt{2} in the combinatorial term underscores the pairwise nature of our comparisons, for instance, between \texttt{t} and \texttt{a1} or \texttt{t} and \texttt{a2}. These comparisons provide insights into the labeling consistency among the various annotators and the true labels.

 Null Hypothesis, $H_0$: There are no differences in population proportion for emo
 among true labels, annotator \texttt{1}, annotator \texttt{2}, annotator \texttt{3}, and ChatGPT
\[P_{t} = P_{a1} = P_{a2} = P_{a3} = P_{g}\]

Alternative Hypothesis, $H_a$: There are differences in population proportion for emo among true labels, annotator \texttt{1}, annotator \texttt{2}, annotator \texttt{3} and ChatGPT \[ P_{t} \neq P_{a1} \neq P_{a2} \neq P_{a3} \neq P_{g} \] where,

Let \(\text{emotion}_1 = \text{anger}\), \(\text{emotion}_2 = \text{cognitive dysfunction}\),
\(\text{emotion}_3 = \text{emptiness}\), \(\text{emotion}_4 = \text{hopelessness}\),
\(\text{emotion}_5 = \text{loneliness}\), \(\text{emotion}_6 = \text{sadness}\),
\(\text{emotion}_7 = \text{suicide intent}\), \(\text{emotion}_8 = \text{worthlessness}\).
Then the sum \(\text{emo}\) can be expressed as:

\[
emo = \sum_{i=1}^{8} \text{emotion}_i
\]

\begin{table*}[!htb]
\small
\centering
\caption{Population proportion analysis (true labels, 3 annotators, and ChatGPT).}

\resizebox{\textwidth}{!}{
\begin{tabular}{llll}
\hline
\textbf{Emotion} & \textbf{Result} & \textbf{Decision} & \textbf{Pairwise Comparison}\\
\hline
anger & 0.832 & FTR H0 & None\\
\hline
cognitive dysfunction & 0.804 & FTR H0 & None\\
\hline
emptiness & 0.814 & FTR H0 & None  \\
\hline
hopelessness & 0.004** & R H0 & (T, a1, a2, a3) and (a3, g) \\
\hline
loneliness & $<$0.001*** & R H0 & (T, a2, a3) and (a1, g)\\
\hline
sadness & 0.804 &  FTR H0 & None \\ 
\hline
suicide intent  & 0.826 & FTR H0 & None \\
\hline
worthlessness & 0.339 &  FTR H0 & None \\
\hline
\multicolumn{4}{l}{\textbf{***}: P value $<$ 0.001, \textbf{**}: P value $<$ 0.05 } \\
\multicolumn{4}{l}{\textbf{FTR}: Fail to reject, \textbf{R}: Reject, \textbf{H0}: null hypothesis } \\
\multicolumn{4}{l}{\textbf{T}: DepressionEmo true labels, \textbf{a1}: annotator 1, \textbf{a2}: annotator 2, \textbf{a3}: annotator 3,\textbf{g}: ChatGPT}
 \\
\multicolumn{4}{l}{each tuple means similarity i.e. (T, a1, a2, a3), (a3, g), (T, a2, a3) and (a1, g) }\\
\hline
\end{tabular}
}
\label{tab:population_proportion}
\end{table*}

\Cref{tab:population_proportion} shows that there is no need to have a pairwise comparison among true labels, \texttt{3} annotators, and ChatGPT except for ``hopelessness'' and ``loneliness''. Hence, the annotations are similar for all emotions except ``hopelessness'' and ``loneliness''. Thus, these two emotions, ``hopelessness'', and ``loneliness'', exhibit discrepancies in their counts, indicating errors. Conducting a population proportion analysis can elucidate whether these discrepancies arise from distinct variations across all five elements (true labels, three annotators, and ChatGPT) or whether there are subsets within this group where several elements share similarities in their counts.

Hopelessness shows significant statistical disparities within two distinct pairs: \texttt{(T, a1, a2, a3)} and \texttt{(a3, g)}. Such findings imply that ChatGPT's annotations are at variance with those of the initial tuple comprising true labels and the first three annotators. Interestingly, annotator \texttt{3} demonstrates congruence with both collectives, signaling a unique interpretation alignment. The term 'tuple' here is used to signify a set of statistically distinguishable elements, as determined by the proportional counts. In the case of ``loneliness'', the distribution of annotations reveals a statistically significant congruence between true labels, annotator \texttt{2}, and annotator \texttt{3}, suggesting these sources concur in their labeling frequency for this emotion. In contrast, annotator \texttt{1} and ChatGPT form another ``tuple'' where the annotation frequency is statistically significant and distinct from the former ``tuple''.

\begin{figure}[!htb]
  \centering
  \includegraphics[width=\linewidth]{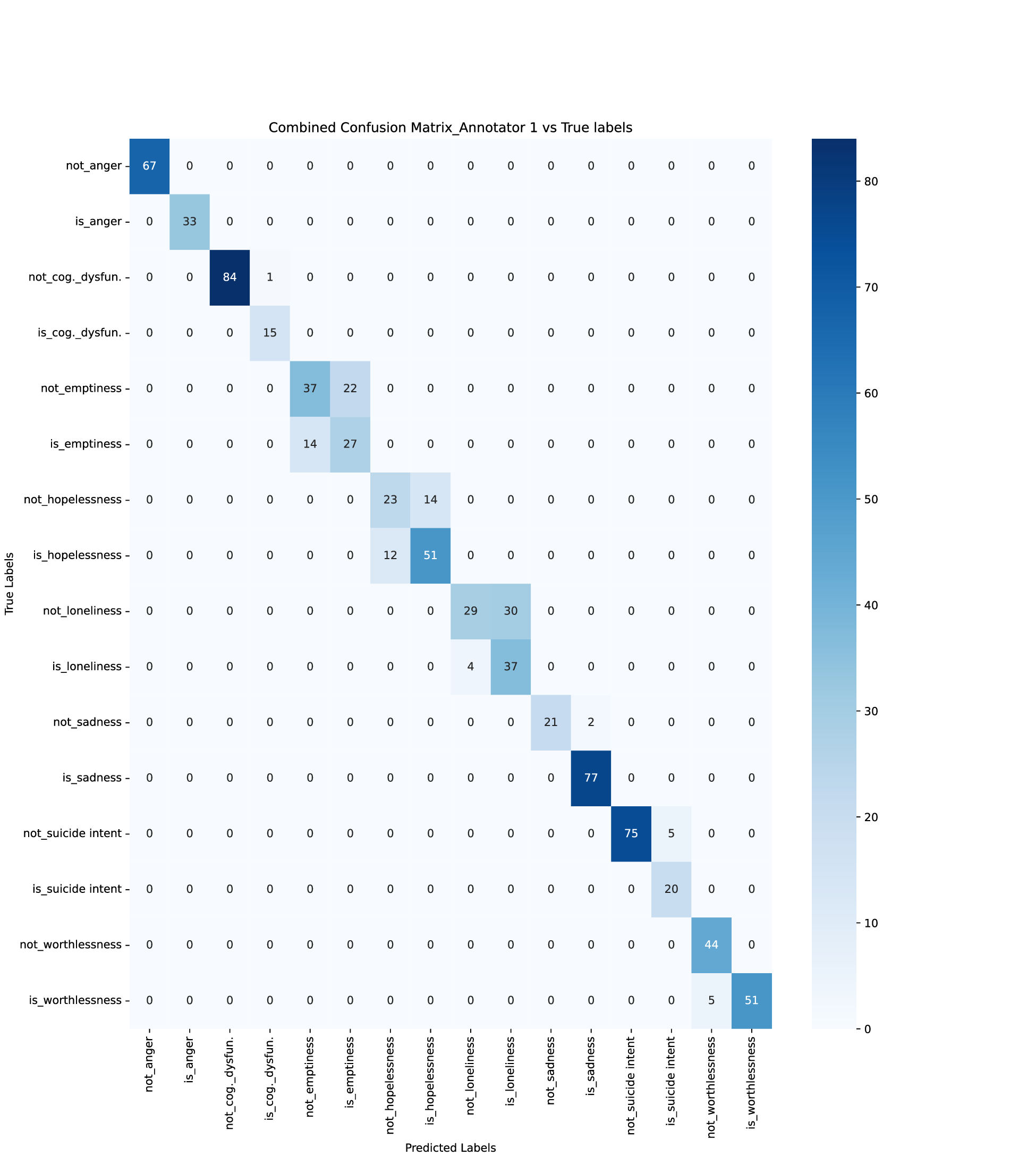}
  \caption{Confusion matrix (annotator 1 vs. true labels)}
  \label{fig:a1}
\end{figure}

\begin{figure}[!htb]
  \centering
  \includegraphics[width=\linewidth]{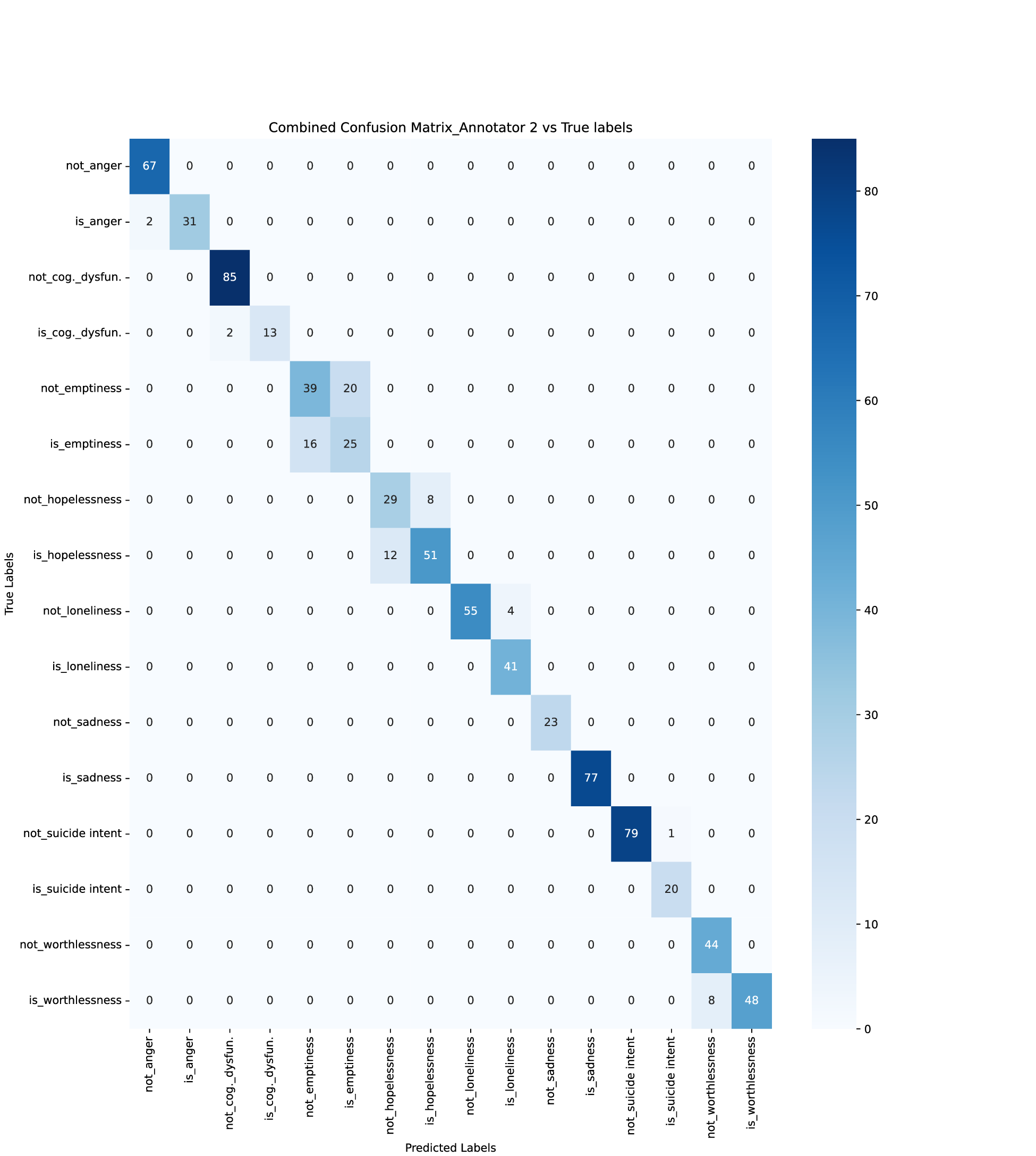}
  \caption{Confusion matrix (annotator 2 vs. true labels)}
  \label{fig:a2}
\end{figure}

\begin{figure}[!htb]
  \centering
  \includegraphics[width=\linewidth]{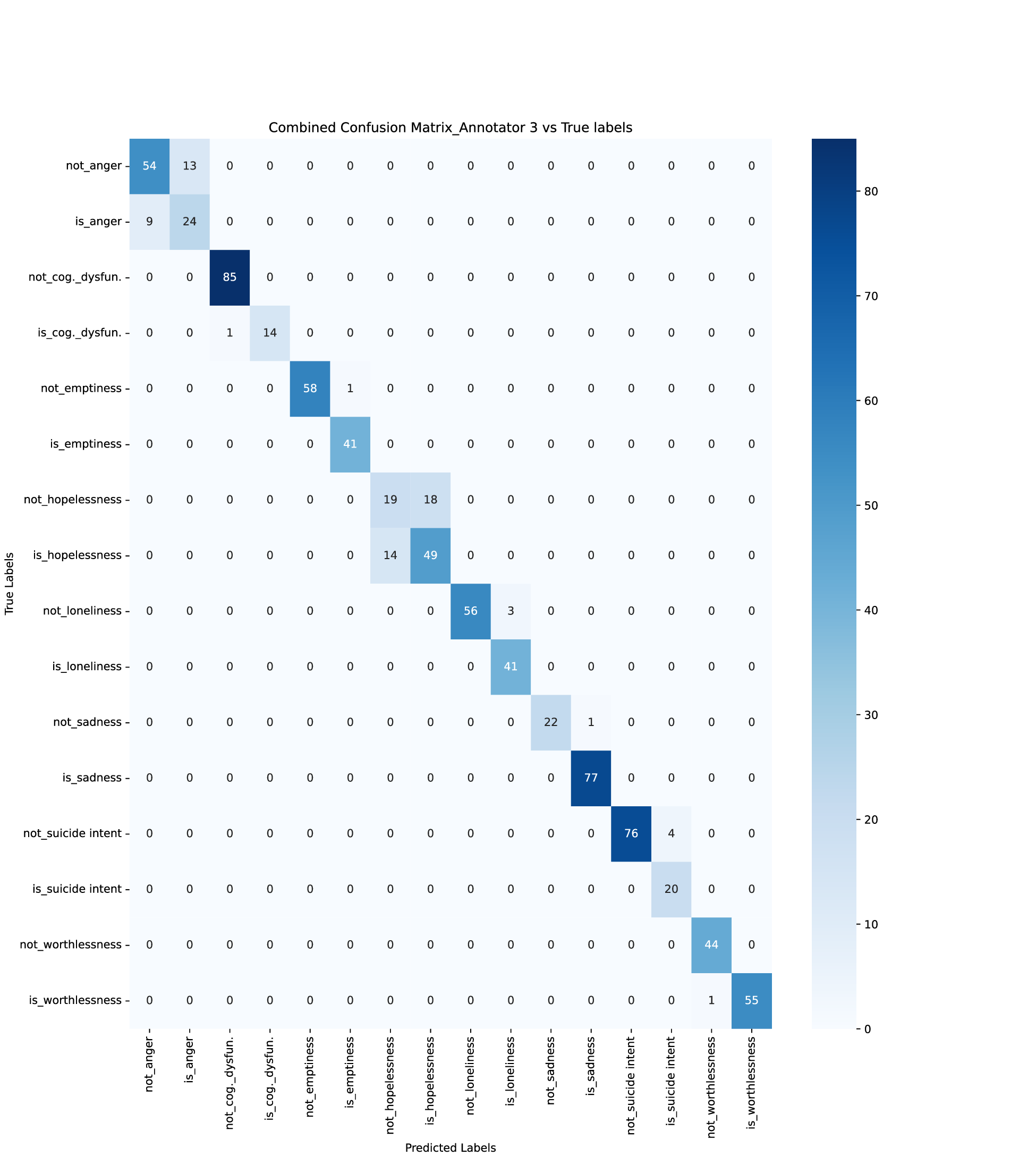}
  \caption{Confusion matrix (annotator 3 vs. true labels)}
  \label{fig:a3}
\end{figure}

\begin{figure}[!htb]
  \centering
  \includegraphics[width=\linewidth]{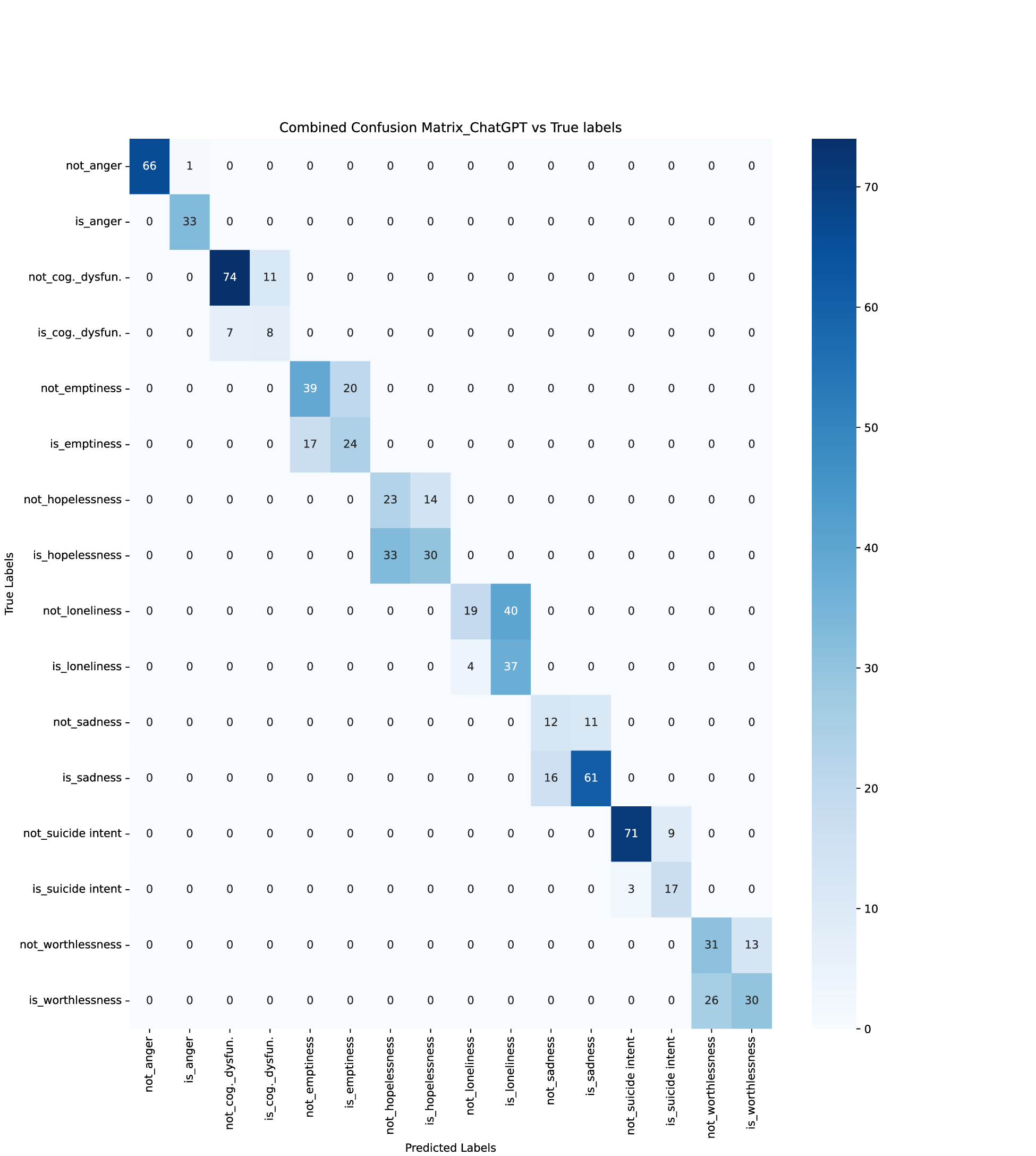}
  \caption{Confusion matrix (ChatGPT vs. true labels)}
  \label{fig:ChatGPT}
\end{figure}

In \Cref{fig:a1}, \Cref{fig:a2}, \Cref{fig:a3}, and \Cref{fig:ChatGPT}, the confusion matrices illustrate the agreement between each annotator and the true labels, as well as the comparison between ChatGPT's labels and the true labels. Within these matrices, ``cognitive dysfunction'' is abbreviated as ``cog.\_dysfun''. These matrices provide a detailed breakdown of \texttt{TP}, \texttt{TN}, \texttt{FP}, and \texttt{FN}, offering valuable insights into the dataset's annotation across human annotators, ChatGPT, and the true labels. Our prior analyses revealed discrepancies in the counts of ``hopelessness'' and ``loneliness'' across the five labeling procedures. Specifically, the confusion matrices facilitate an examination of:

\begin{enumerate}
    \item The labeling differences for ``hopelessness'' between Annotator \texttt{3 (a3)} and ChatGPT \texttt{(g)} compared to the true labels \texttt{(t)}. For instance, in the \texttt{a3 vs. t} comparison, there are \texttt{49 TP}, \texttt{19 TN}, \texttt{14 FP}, and \texttt{18 FN} for ``hopelessness''.
    \item The labeling differences for ``loneliness'' between Annotator \texttt{1 (a1)} and ChatGPT \texttt{(g)} compared to the true labels \texttt{(t)}, where \texttt{TPs}, \texttt{FPs}, \texttt{FNs}, and \texttt{TNs} are \texttt{37}, \texttt{4}, \texttt{30}, and \texttt{29}, respectively.
\end{enumerate}

More broadly, the confusion matrices delineate subtle discrepancies in labeling for each emotion across all annotators and ChatGPT relative to the true labels. For example, analyzing the ``anger'' emotion reveals that no pairwise comparison is necessary among the five labeling categories. Now, based on ``is\_anger'' from the confusion matrix, annotator \texttt{1} identified ``anger'' \texttt{33} times, while Annotators \texttt{2} and \texttt{3}, along with the true labels and ChatGPT, counted it \texttt{31}, \texttt{37}, \texttt{38}, and \texttt{34} times, respectively. The similarity in counts between Annotator \texttt{3} and the true labels (\texttt{37} and \texttt{38}, respectively) suggests a high level of agreement in identifying ``anger''.

However, the confusion matrix for Annotator \texttt{3} vs. the true labels, with \texttt{24 TP}, \texttt{54 TN}, \texttt{9 FP}, and \texttt{13 FN}, reveals nuanced differences. To illustrate, consider a scenario with a random selection of five samples where Annotator \texttt{3} labels them as \texttt{0}, \texttt{1}, \texttt{0}, \texttt{1}, \texttt{0}, and the true labels are \texttt{1}, \texttt{0}, \texttt{0}, \texttt{1}, \texttt{0}. While the total counts of ``not\_anger'' \texttt{(0)} and ``is\_anger'' \texttt{(1)} match; their allocation across specific samples varies. Therefore, while the population proportion analysis shows no significant difference, the confusion matrix provides a deeper understanding of the labeling process.

Although there is a high degree of agreement between the annotators and ChatGPT in the dataset's labeling, our error analysis highlights certain instances of minor discrepancies.

\section{Limitation}
Compared to other available datasets, \dataset{} is comparatively small, particularly when addressing the challenge of multilabel classification involving 8 labels. As a result, \dataset{} may not encompass all possible emotion combinations and capture the sample diversity well. However, we believe that this limitation could be addressed by gathering additional data.

The second limitation is that the annotation arises from the constraints of human resources and the associated costs. Increased perfection in annotation could be achieved by involving more annotators and allowing each annotator to label the data twice or thrice. This approach helps mitigate potential typos that may occur during the coding process, and it also leverages the annotator's insights to refine the annotations through multiple iterations.

\section{Conclusion}
In this paper, we created a novel dataset named \dataset{} designed for identifying depression-related emotions from Reddit user posts. We also presented details about the data collection process and assessed annotation quality by evaluating inter-rater reliability, ensuring it met acceptable standards. Subsequently, we conducted various analyses on \dataset{}, examining text length distribution and emotion distribution, with a specific focus on the temporal evolution of suicide intent. 

Our approach involved employing established text classification methods and configuring a single classifier to detect all emotions simultaneously. The experimental findings indicated the superiority of deep learning methods over machine learning ones in text classification, particularly in capturing semantic meanings. Notably, BART demonstrated superior performance compared to other methods, with BERT closely trailing. Besides emotion identification, \dataset{} proved valuable in identifying emotions among individuals exhibiting symptoms of depression in texts, with suicide intent yielding the highest F1-Macro value.

In the future, our work plan involves enlarging the dataset by adding more examples and encompassing a wider spectrum of emotions. We intend to improve the dataset's quality by incorporating a blend of human annotations and annotations generated through ChatGPT-based methods, as highlighted in previous works~\cite{zhu2023can,gilardi2023chatgpt,belal2023leveraging}. Furthermore, we expect to uncover more efficient and novel text classification approaches, especially in the context of large language models. Ultimately, we expect our paper to become a valuable reference for precisely detecting indicators of depression in textual content.

\section*{Data Availability Statement}
\newcommand*{\brokenurlwithoutpar}[2]{{\texttt{#1}}\\*{\texttt{#2}}}
The curated dataset is publicly available at:
\begin{itemize}
    \item \url{https://github.com/abuBakarSiddiqurRahman/DepressionEmo}
\end{itemize}

\section*{Compliance with Ethical Standards}
\begin{itemize}
    \item This article does not contain any studies with human participants or animals performed by any of the authors.
    \item All authors certify that they have no affiliations with or involvement in any organization or entity with any financial interest or non-financial interest in the subject matter or materials discussed in this manuscript.
\end{itemize}

 






\bibliographystyle{elsarticle-num-names}
\bibliography{elsarticle-template-num-names}

\begin{thebibliography}{75}
\expandafter\ifx\csname natexlab\endcsname\relax\def\natexlab#1{#1}\fi
\providecommand{\url}[1]{\texttt{#1}}
\providecommand{\href}[2]{#2}
\providecommand{\path}[1]{#1}
\providecommand{\DOIprefix}{doi:}
\providecommand{\ArXivprefix}{arXiv:}
\providecommand{\URLprefix}{URL: }
\providecommand{\Pubmedprefix}{pmid:}
\providecommand{\doi}[1]{\href{http://dx.doi.org/#1}{\path{#1}}}
\providecommand{\Pubmed}[1]{\href{pmid:#1}{\path{#1}}}
\providecommand{\bibinfo}[2]{#2}
\ifx\xfnm\relax \def\xfnm[#1]{\unskip,\space#1}\fi
\bibitem[{Zhang et~al.(2019)Zhang, Driscol, Chen, and Hosseini~Ghomi}]{zhang2019evaluating}
\bibinfo{author}{L.~Zhang}, \bibinfo{author}{J.~Driscol}, \bibinfo{author}{X.~Chen}, \bibinfo{author}{R.~Hosseini~Ghomi},
\newblock \bibinfo{title}{Evaluating acoustic and linguistic features of detecting depression sub-challenge dataset},
\newblock in: \bibinfo{booktitle}{Proceedings of the 9th International on Audio/Visual Emotion Challenge and Workshop}, \bibinfo{year}{2019}, pp. \bibinfo{pages}{47--53}.
\bibitem[{Morrill et~al.(2008)Morrill, Brewer, O'Neill, Lillie, Dees, Carey, and Rimer}]{morrill2008interaction}
\bibinfo{author}{E.~F. Morrill}, \bibinfo{author}{N.~T. Brewer}, \bibinfo{author}{S.~C. O'Neill}, \bibinfo{author}{S.~E. Lillie}, \bibinfo{author}{E.~C. Dees}, \bibinfo{author}{L.~A. Carey}, \bibinfo{author}{B.~K. Rimer},
\newblock \bibinfo{title}{The interaction of post-traumatic growth and post-traumatic stress symptoms in predicting depressive symptoms and quality of life},
\newblock \bibinfo{journal}{Psycho-Oncology: Journal of the Psychological, Social and Behavioral Dimensions of Cancer} \bibinfo{volume}{17} (\bibinfo{year}{2008}) \bibinfo{pages}{948--953}.
\bibitem[{Ren et~al.(2021)Ren, Lin, Xu, Zhang, Yang, and Sun}]{Ren2021Depression}
\bibinfo{author}{L.~Ren}, \bibinfo{author}{H.~Lin}, \bibinfo{author}{B.~Xu}, \bibinfo{author}{S.~Zhang}, \bibinfo{author}{L.~Yang}, \bibinfo{author}{S.~Sun},
\newblock \bibinfo{title}{Depression detection on reddit with an emotion-based attention network: algorithm development and validation},
\newblock \bibinfo{journal}{JMIR medical informatics} \bibinfo{volume}{9} (\bibinfo{year}{2021}) \bibinfo{pages}{e28754}.
\bibitem[{Guo et~al.(2023)Guo, Liu, Wang, Qin, Hao, and Hong}]{guo2023prompt}
\bibinfo{author}{Y.~Guo}, \bibinfo{author}{J.~Liu}, \bibinfo{author}{L.~Wang}, \bibinfo{author}{W.~Qin}, \bibinfo{author}{S.~Hao}, \bibinfo{author}{R.~Hong},
\newblock \bibinfo{title}{A prompt-based topic-modeling method for depression detection on low-resource data},
\newblock \bibinfo{journal}{IEEE Transactions on Computational Social Systems}  (\bibinfo{year}{2023}).
\bibitem[{Wang et~al.(2023)Wang, Zhao, Lu, and Qin}]{wang2023cognitive}
\bibinfo{author}{B.~Wang}, \bibinfo{author}{Y.~Zhao}, \bibinfo{author}{X.~Lu}, \bibinfo{author}{B.~Qin},
\newblock \bibinfo{title}{Cognitive distortion based explainable depression detection and analysis technologies for the adolescent internet users on social media},
\newblock \bibinfo{journal}{Frontiers in Public Health} \bibinfo{volume}{10} (\bibinfo{year}{2023}) \bibinfo{pages}{1045777}.
\bibitem[{Lyu et~al.(2023)Lyu, Ren, Du, and Zhao}]{lyu2023detecting}
\bibinfo{author}{S.~Lyu}, \bibinfo{author}{X.~Ren}, \bibinfo{author}{Y.~Du}, \bibinfo{author}{N.~Zhao},
\newblock \bibinfo{title}{Detecting depression of chinese microblog users via text analysis: Combining linguistic inquiry word count (liwc) with culture and suicide related lexicons},
\newblock \bibinfo{journal}{Frontiers in psychiatry} \bibinfo{volume}{14} (\bibinfo{year}{2023}) \bibinfo{pages}{1121583}.
\bibitem[{Zhang et~al.(2023)Zhang, Xie, Liu, and Zhang}]{zhang2023depression}
\bibinfo{author}{W.~Zhang}, \bibinfo{author}{J.~Xie}, \bibinfo{author}{X.~Liu}, \bibinfo{author}{Z.~Zhang},
\newblock \bibinfo{title}{Depression detection using digital traces on social media: A knowledge-aware deep learning approach},
\newblock \bibinfo{journal}{arXiv e-prints}  (\bibinfo{year}{2023}) \bibinfo{pages}{arXiv--2303}.
\bibitem[{De~Melo et~al.(2019)De~Melo, Granger, and Hadid}]{de2019depression}
\bibinfo{author}{W.~C. De~Melo}, \bibinfo{author}{E.~Granger}, \bibinfo{author}{A.~Hadid},
\newblock \bibinfo{title}{Depression detection based on deep distribution learning},
\newblock in: \bibinfo{booktitle}{2019 IEEE international conference on image processing (ICIP)}, \bibinfo{organization}{IEEE}, \bibinfo{year}{2019}, pp. \bibinfo{pages}{4544--4548}.
\bibitem[{Han et~al.(2022)Han, Mao, and Cambria}]{han2022hierarchical}
\bibinfo{author}{S.~Han}, \bibinfo{author}{R.~Mao}, \bibinfo{author}{E.~Cambria},
\newblock \bibinfo{title}{Hierarchical attention network for explainable depression detection on twitter aided by metaphor concept mappings},
\newblock \bibinfo{journal}{arXiv preprint arXiv:2209.07494}  (\bibinfo{year}{2022}).
\bibitem[{Wang et~al.(2020)Wang, Chen, Li, Li, Zhou, Zheng, Chen, Yan, Tang et~al.}]{wang2020depression}
\bibinfo{author}{X.~Wang}, \bibinfo{author}{S.~Chen}, \bibinfo{author}{T.~Li}, \bibinfo{author}{W.~Li}, \bibinfo{author}{Y.~Zhou}, \bibinfo{author}{J.~Zheng}, \bibinfo{author}{Q.~Chen}, \bibinfo{author}{J.~Yan}, \bibinfo{author}{B.~Tang}, et~al.,
\newblock \bibinfo{title}{Depression risk prediction for chinese microblogs via deep-learning methods: Content analysis},
\newblock \bibinfo{journal}{JMIR medical informatics} \bibinfo{volume}{8} (\bibinfo{year}{2020}) \bibinfo{pages}{e17958}.
\bibitem[{Heston(2023)}]{heston2023safety}
\bibinfo{author}{T.~F. Heston},
\newblock \bibinfo{title}{Safety of large language models in addressing depression},
\newblock \bibinfo{journal}{Cureus} \bibinfo{volume}{15} (\bibinfo{year}{2023}).
\bibitem[{Xu et~al.(2023)Xu, Yao, Dong, Yu, Hendler, Dey, and Wang}]{xu2023leveraging}
\bibinfo{author}{X.~Xu}, \bibinfo{author}{B.~Yao}, \bibinfo{author}{Y.~Dong}, \bibinfo{author}{H.~Yu}, \bibinfo{author}{J.~Hendler}, \bibinfo{author}{A.~K. Dey}, \bibinfo{author}{D.~Wang},
\newblock \bibinfo{title}{Leveraging large language models for mental health prediction via online text data},
\newblock \bibinfo{journal}{arXiv preprint arXiv:2307.14385}  (\bibinfo{year}{2023}).
\bibitem[{Lamichhane(2023)}]{lamichhane2023evaluation}
\bibinfo{author}{B.~Lamichhane},
\newblock \bibinfo{title}{Evaluation of chatgpt for nlp-based mental health applications},
\newblock \bibinfo{journal}{arXiv preprint arXiv:2303.15727}  (\bibinfo{year}{2023}).
\bibitem[{Chiong et~al.(2021)Chiong, Budhi, Dhakal, and Chiong}]{chiong2021textual}
\bibinfo{author}{R.~Chiong}, \bibinfo{author}{G.~S. Budhi}, \bibinfo{author}{S.~Dhakal}, \bibinfo{author}{F.~Chiong},
\newblock \bibinfo{title}{A textual-based featuring approach for depression detection using machine learning classifiers and social media texts},
\newblock \bibinfo{journal}{Computers in Biology and Medicine} \bibinfo{volume}{135} (\bibinfo{year}{2021}) \bibinfo{pages}{104499}.
\bibitem[{Al~Asad et~al.(2019)Al~Asad, Pranto, Afreen, and Islam}]{al2019depression}
\bibinfo{author}{N.~Al~Asad}, \bibinfo{author}{M.~A.~M. Pranto}, \bibinfo{author}{S.~Afreen}, \bibinfo{author}{M.~M. Islam},
\newblock \bibinfo{title}{Depression detection by analyzing social media posts of user},
\newblock in: \bibinfo{booktitle}{2019 IEEE international conference on signal processing, information, communication \& systems (SPICSCON)}, \bibinfo{organization}{IEEE}, \bibinfo{year}{2019}, pp. \bibinfo{pages}{13--17}.
\bibitem[{Katchapakirin et~al.(2018)Katchapakirin, Wongpatikaseree, Yomaboot, and Kaewpitakkun}]{katchapakirin2018facebook}
\bibinfo{author}{K.~Katchapakirin}, \bibinfo{author}{K.~Wongpatikaseree}, \bibinfo{author}{P.~Yomaboot}, \bibinfo{author}{Y.~Kaewpitakkun},
\newblock \bibinfo{title}{Facebook social media for depression detection in the thai community},
\newblock in: \bibinfo{booktitle}{2018 15th international joint conference on computer science and software engineering (jcsse)}, \bibinfo{organization}{IEEE}, \bibinfo{year}{2018}, pp. \bibinfo{pages}{1--6}.
\bibitem[{Eichstaedt et~al.(2018)Eichstaedt, Smith, Merchant, Ungar, Crutchley, Preo{\c{t}}iuc-Pietro, Asch, and Schwartz}]{eichstaedt2018facebook}
\bibinfo{author}{J.~C. Eichstaedt}, \bibinfo{author}{R.~J. Smith}, \bibinfo{author}{R.~M. Merchant}, \bibinfo{author}{L.~H. Ungar}, \bibinfo{author}{P.~Crutchley}, \bibinfo{author}{D.~Preo{\c{t}}iuc-Pietro}, \bibinfo{author}{D.~A. Asch}, \bibinfo{author}{H.~A. Schwartz},
\newblock \bibinfo{title}{Facebook language predicts depression in medical records},
\newblock \bibinfo{journal}{Proceedings of the National Academy of Sciences} \bibinfo{volume}{115} (\bibinfo{year}{2018}) \bibinfo{pages}{11203--11208}.
\bibitem[{Chen et~al.(2023)Chen, Yang, Fu, Zong, Liu, and Huang}]{chen2023detecting}
\bibinfo{author}{Z.~Chen}, \bibinfo{author}{R.~Yang}, \bibinfo{author}{S.~Fu}, \bibinfo{author}{N.~Zong}, \bibinfo{author}{H.~Liu}, \bibinfo{author}{M.~Huang},
\newblock \bibinfo{title}{Detecting reddit users with depression using a hybrid neural network},
\newblock \bibinfo{journal}{arXiv preprint arXiv:2302.02759}  (\bibinfo{year}{2023}).
\bibitem[{Nguyen et~al.(2014)Nguyen, Phung, Dao, Venkatesh, and Berk}]{nguyen2014affective}
\bibinfo{author}{T.~Nguyen}, \bibinfo{author}{D.~Phung}, \bibinfo{author}{B.~Dao}, \bibinfo{author}{S.~Venkatesh}, \bibinfo{author}{M.~Berk},
\newblock \bibinfo{title}{Affective and content analysis of online depression communities},
\newblock \bibinfo{journal}{IEEE transactions on affective computing} \bibinfo{volume}{5} (\bibinfo{year}{2014}) \bibinfo{pages}{217--226}.
\bibitem[{Chen et~al.(2015)Chen, He, Benesty, Khotilovich, Tang, Cho, Chen, Mitchell, Cano, Zhou et~al.}]{chen2015xgboost}
\bibinfo{author}{T.~Chen}, \bibinfo{author}{T.~He}, \bibinfo{author}{M.~Benesty}, \bibinfo{author}{V.~Khotilovich}, \bibinfo{author}{Y.~Tang}, \bibinfo{author}{H.~Cho}, \bibinfo{author}{K.~Chen}, \bibinfo{author}{R.~Mitchell}, \bibinfo{author}{I.~Cano}, \bibinfo{author}{T.~Zhou}, et~al.,
\newblock \bibinfo{title}{Xgboost: extreme gradient boosting},
\newblock \bibinfo{journal}{R package version 0.4-2} \bibinfo{volume}{1} (\bibinfo{year}{2015}) \bibinfo{pages}{1--4}.
\bibitem[{Ke et~al.(2017)Ke, Meng, Finley, Wang, Chen, Ma, Ye, and Liu}]{ke2017lightgbm}
\bibinfo{author}{G.~Ke}, \bibinfo{author}{Q.~Meng}, \bibinfo{author}{T.~Finley}, \bibinfo{author}{T.~Wang}, \bibinfo{author}{W.~Chen}, \bibinfo{author}{W.~Ma}, \bibinfo{author}{Q.~Ye}, \bibinfo{author}{T.-Y. Liu},
\newblock \bibinfo{title}{Lightgbm: A highly efficient gradient boosting decision tree},
\newblock \bibinfo{journal}{Advances in neural information processing systems} \bibinfo{volume}{30} (\bibinfo{year}{2017}).
\bibitem[{Devlin et~al.(2018)Devlin, Chang, Lee, and Toutanova}]{devlin2018bert}
\bibinfo{author}{J.~Devlin}, \bibinfo{author}{M.-W. Chang}, \bibinfo{author}{K.~Lee}, \bibinfo{author}{K.~Toutanova},
\newblock \bibinfo{title}{Bert: Pre-training of deep bidirectional transformers for language understanding},
\newblock \bibinfo{journal}{arXiv preprint arXiv:1810.04805}  (\bibinfo{year}{2018}).
\bibitem[{Croce et~al.(2020)Croce, Castellucci, and Basili}]{croce2020gan}
\bibinfo{author}{D.~Croce}, \bibinfo{author}{G.~Castellucci}, \bibinfo{author}{R.~Basili},
\newblock \bibinfo{title}{Gan-bert: Generative adversarial learning for robust text classification with a bunch of labeled examples}  (\bibinfo{year}{2020}).
\bibitem[{Lewis et~al.(2019)Lewis, Liu, Goyal, Ghazvininejad, Mohamed, Levy, Stoyanov, and Zettlemoyer}]{lewis2019bart}
\bibinfo{author}{M.~Lewis}, \bibinfo{author}{Y.~Liu}, \bibinfo{author}{N.~Goyal}, \bibinfo{author}{M.~Ghazvininejad}, \bibinfo{author}{A.~Mohamed}, \bibinfo{author}{O.~Levy}, \bibinfo{author}{V.~Stoyanov}, \bibinfo{author}{L.~Zettlemoyer},
\newblock \bibinfo{title}{Bart: Denoising sequence-to-sequence pre-training for natural language generation, translation, and comprehension},
\newblock \bibinfo{journal}{arXiv preprint arXiv:1910.13461}  (\bibinfo{year}{2019}).
\bibitem[{Hasin et~al.(2005)Hasin, Goodwin, Stinson, and Grant}]{hasin2005epidemiology}
\bibinfo{author}{D.~S. Hasin}, \bibinfo{author}{R.~D. Goodwin}, \bibinfo{author}{F.~S. Stinson}, \bibinfo{author}{B.~F. Grant},
\newblock \bibinfo{title}{Epidemiology of major depressive disorder: results from the national epidemiologic survey on alcoholism and related conditions},
\newblock \bibinfo{journal}{Archives of general psychiatry} \bibinfo{volume}{62} (\bibinfo{year}{2005}) \bibinfo{pages}{1097--1106}.
\bibitem[{Weinberger et~al.(2018)Weinberger, Gbedemah, Martinez, Nash, Galea, and Goodwin}]{weinberger2018trends}
\bibinfo{author}{A.~H. Weinberger}, \bibinfo{author}{M.~Gbedemah}, \bibinfo{author}{A.~M. Martinez}, \bibinfo{author}{D.~Nash}, \bibinfo{author}{S.~Galea}, \bibinfo{author}{R.~D. Goodwin},
\newblock \bibinfo{title}{Trends in depression prevalence in the usa from 2005 to 2015: widening disparities in vulnerable groups},
\newblock \bibinfo{journal}{Psychological medicine} \bibinfo{volume}{48} (\bibinfo{year}{2018}) \bibinfo{pages}{1308--1315}.
\bibitem[{Menselson et~al.(2008)Menselson, Rehkopf, and Kubzansky}]{menselson2008depression}
\bibinfo{author}{T.~Menselson}, \bibinfo{author}{D.~H. Rehkopf}, \bibinfo{author}{L.~D. Kubzansky},
\newblock \bibinfo{title}{Depression among latinos in the united states: a meta-analytic review.},
\newblock \bibinfo{journal}{Journal of Consulting and Clinical Psychology} \bibinfo{volume}{76} (\bibinfo{year}{2008}) \bibinfo{pages}{355}.
\bibitem[{Berry and York(2011)}]{berry2011depression}
\bibinfo{author}{D.~M. Berry}, \bibinfo{author}{K.~York},
\newblock \bibinfo{title}{Depression and religiosity and/or spirituality in college: A longitudinal survey of students in the usa},
\newblock \bibinfo{journal}{Nursing \& health sciences} \bibinfo{volume}{13} (\bibinfo{year}{2011}) \bibinfo{pages}{76--83}.
\bibitem[{Olfson et~al.(2016)Olfson, Blanco, and Marcus}]{olfson2016treatment}
\bibinfo{author}{M.~Olfson}, \bibinfo{author}{C.~Blanco}, \bibinfo{author}{S.~C. Marcus},
\newblock \bibinfo{title}{Treatment of adult depression in the united states},
\newblock \bibinfo{journal}{JAMA internal medicine} \bibinfo{volume}{176} (\bibinfo{year}{2016}) \bibinfo{pages}{1482--1491}.
\bibitem[{Cong et~al.(2018)Cong, Feng, Li, Xiang, Rao, and Tao}]{cong2018xa}
\bibinfo{author}{Q.~Cong}, \bibinfo{author}{Z.~Feng}, \bibinfo{author}{F.~Li}, \bibinfo{author}{Y.~Xiang}, \bibinfo{author}{G.~Rao}, \bibinfo{author}{C.~Tao},
\newblock \bibinfo{title}{Xa-bilstm: a deep learning approach for depression detection in imbalanced data},
\newblock in: \bibinfo{booktitle}{2018 IEEE international conference on bioinformatics and biomedicine (BIBM)}, \bibinfo{organization}{IEEE}, \bibinfo{year}{2018}, pp. \bibinfo{pages}{1624--1627}.
\bibitem[{Zogan et~al.(2022)Zogan, Razzak, Wang, Jameel, and Xu}]{zogan2022explainable}
\bibinfo{author}{H.~Zogan}, \bibinfo{author}{I.~Razzak}, \bibinfo{author}{X.~Wang}, \bibinfo{author}{S.~Jameel}, \bibinfo{author}{G.~Xu},
\newblock \bibinfo{title}{Explainable depression detection with multi-aspect features using a hybrid deep learning model on social media},
\newblock \bibinfo{journal}{World Wide Web} \bibinfo{volume}{25} (\bibinfo{year}{2022}) \bibinfo{pages}{281--304}.
\bibitem[{Tavchioski et~al.(2023)Tavchioski, Robnik-{\v{S}}ikonja, and Pollak}]{tavchioski2023detection}
\bibinfo{author}{I.~Tavchioski}, \bibinfo{author}{M.~Robnik-{\v{S}}ikonja}, \bibinfo{author}{S.~Pollak},
\newblock \bibinfo{title}{Detection of depression on social networks using transformers and ensembles},
\newblock \bibinfo{journal}{arXiv preprint arXiv:2305.05325}  (\bibinfo{year}{2023}).
\bibitem[{Kabir et~al.(2023)Kabir, Ahmed, Hasan, Laskar, Joarder, Mahmud, and Hasan}]{kabir2023deptweet}
\bibinfo{author}{M.~Kabir}, \bibinfo{author}{T.~Ahmed}, \bibinfo{author}{M.~B. Hasan}, \bibinfo{author}{M.~T.~R. Laskar}, \bibinfo{author}{T.~K. Joarder}, \bibinfo{author}{H.~Mahmud}, \bibinfo{author}{K.~Hasan},
\newblock \bibinfo{title}{Deptweet: A typology for social media texts to detect depression severities},
\newblock \bibinfo{journal}{Computers in Human Behavior} \bibinfo{volume}{139} (\bibinfo{year}{2023}) \bibinfo{pages}{107503}.
\bibitem[{Stepanov et~al.(2018)Stepanov, Lathuiliere, Chowdhury, Ghosh, Vieriu, Sebe, and Riccardi}]{stepanov2018depression}
\bibinfo{author}{E.~A. Stepanov}, \bibinfo{author}{S.~Lathuiliere}, \bibinfo{author}{S.~A. Chowdhury}, \bibinfo{author}{A.~Ghosh}, \bibinfo{author}{R.-L. Vieriu}, \bibinfo{author}{N.~Sebe}, \bibinfo{author}{G.~Riccardi},
\newblock \bibinfo{title}{Depression severity estimation from multiple modalities},
\newblock in: \bibinfo{booktitle}{2018 ieee 20th international conference on e-health networking, applications and services (healthcom)}, \bibinfo{organization}{IEEE}, \bibinfo{year}{2018}, pp. \bibinfo{pages}{1--6}.
\bibitem[{Trotzek et~al.(2018)Trotzek, Koitka, and Friedrich}]{trotzek2018utilizing}
\bibinfo{author}{M.~Trotzek}, \bibinfo{author}{S.~Koitka}, \bibinfo{author}{C.~M. Friedrich},
\newblock \bibinfo{title}{Utilizing neural networks and linguistic metadata for early detection of depression indications in text sequences},
\newblock \bibinfo{journal}{IEEE Transactions on Knowledge and Data Engineering} \bibinfo{volume}{32} (\bibinfo{year}{2018}) \bibinfo{pages}{588--601}.
\bibitem[{Mart{\i}nez-Castano et~al.(2020)Mart{\i}nez-Castano, Htait, Azzopardi, and Moshfeghi}]{martinez2020early}
\bibinfo{author}{R.~Mart{\i}nez-Castano}, \bibinfo{author}{A.~Htait}, \bibinfo{author}{L.~Azzopardi}, \bibinfo{author}{Y.~Moshfeghi},
\newblock \bibinfo{title}{Early risk detection of self-harm and depression severity using bert-based transformers},
\newblock \bibinfo{journal}{Working Notes of CLEF}  (\bibinfo{year}{2020}) \bibinfo{pages}{16}.
\bibitem[{DeSouza et~al.(2021)DeSouza, Robin, Gumus, and Yeung}]{desouza2021natural}
\bibinfo{author}{D.~D. DeSouza}, \bibinfo{author}{J.~Robin}, \bibinfo{author}{M.~Gumus}, \bibinfo{author}{A.~Yeung},
\newblock \bibinfo{title}{Natural language processing as an emerging tool to detect late-life depression},
\newblock \bibinfo{journal}{Frontiers in Psychiatry} \bibinfo{volume}{12} (\bibinfo{year}{2021}) \bibinfo{pages}{719125}.
\bibitem[{Zogan et~al.(2021)Zogan, Razzak, Jameel, and Xu}]{zogan2021depressionnet}
\bibinfo{author}{H.~Zogan}, \bibinfo{author}{I.~Razzak}, \bibinfo{author}{S.~Jameel}, \bibinfo{author}{G.~Xu},
\newblock \bibinfo{title}{Depressionnet: A novel summarization boosted deep framework for depression detection on social media},
\newblock \bibinfo{journal}{arXiv preprint arXiv:2105.10878}  (\bibinfo{year}{2021}).
\bibitem[{William et~al.(2022)William, Achmad, Suhartono, and Gema}]{william2022leveraging}
\bibinfo{author}{D.~William}, \bibinfo{author}{S.~Achmad}, \bibinfo{author}{D.~Suhartono}, \bibinfo{author}{A.~P. Gema},
\newblock \bibinfo{title}{Leveraging bert with extractive summarization for depression detection on social media},
\newblock in: \bibinfo{booktitle}{2022 International Seminar on Intelligent Technology and Its Applications (ISITIA)}, \bibinfo{organization}{IEEE}, \bibinfo{year}{2022}, pp. \bibinfo{pages}{63--68}.
\bibitem[{Shen et~al.(2017)Shen, Jia, Nie, Feng, Zhang, Hu, Chua, Zhu et~al.}]{shen2017depression}
\bibinfo{author}{G.~Shen}, \bibinfo{author}{J.~Jia}, \bibinfo{author}{L.~Nie}, \bibinfo{author}{F.~Feng}, \bibinfo{author}{C.~Zhang}, \bibinfo{author}{T.~Hu}, \bibinfo{author}{T.-S. Chua}, \bibinfo{author}{W.~Zhu}, et~al.,
\newblock \bibinfo{title}{Depression detection via harvesting social media: A multimodal dictionary learning solution.},
\newblock in: \bibinfo{booktitle}{IJCAI}, \bibinfo{year}{2017}, pp. \bibinfo{pages}{3838--3844}.
\bibitem[{Coppersmith et~al.(2015)Coppersmith, Dredze, Harman, Hollingshead, and Mitchell}]{coppersmith2015clpsych}
\bibinfo{author}{G.~Coppersmith}, \bibinfo{author}{M.~Dredze}, \bibinfo{author}{C.~Harman}, \bibinfo{author}{K.~Hollingshead}, \bibinfo{author}{M.~Mitchell},
\newblock \bibinfo{title}{Clpsych 2015 shared task: Depression and ptsd on twitter},
\newblock in: \bibinfo{booktitle}{Proceedings of the 2nd workshop on computational linguistics and clinical psychology: from linguistic signal to clinical reality}, \bibinfo{year}{2015}, pp. \bibinfo{pages}{31--39}.
\bibitem[{Skaik and Inkpen(2020)}]{skaik2020using}
\bibinfo{author}{R.~Skaik}, \bibinfo{author}{D.~Inkpen},
\newblock \bibinfo{title}{Using twitter social media for depression detection in the canadian population},
\newblock in: \bibinfo{booktitle}{Proceedings of the 2020 3rd Artificial Intelligence and Cloud Computing Conference}, \bibinfo{year}{2020}, pp. \bibinfo{pages}{109--114}.
\bibitem[{Owen et~al.(2020)Owen, Collados, and Espinosa-Anke}]{owen2020towards}
\bibinfo{author}{D.~Owen}, \bibinfo{author}{J.~C. Collados}, \bibinfo{author}{L.~Espinosa-Anke},
\newblock \bibinfo{title}{Towards preemptive detection of depression and anxiety in twitter},
\newblock \bibinfo{journal}{arXiv preprint arXiv:2011.05249}  (\bibinfo{year}{2020}).
\bibitem[{Pirina and {\c{C}}{\"o}ltekin(2018)}]{pirina2018identifying}
\bibinfo{author}{I.~Pirina}, \bibinfo{author}{{\c{C}}.~{\c{C}}{\"o}ltekin},
\newblock \bibinfo{title}{Identifying depression on reddit: The effect of training data},
\newblock in: \bibinfo{booktitle}{Proceedings of the 2018 EMNLP workshop SMM4H: the 3rd social media mining for health applications workshop \& shared task}, \bibinfo{year}{2018}, pp. \bibinfo{pages}{9--12}.
\bibitem[{Losada and Crestani(2016)}]{losada2016test}
\bibinfo{author}{D.~E. Losada}, \bibinfo{author}{F.~Crestani},
\newblock \bibinfo{title}{A test collection for research on depression and language use},
\newblock in: \bibinfo{booktitle}{International conference of the cross-language evaluation forum for European languages}, \bibinfo{organization}{Springer}, \bibinfo{year}{2016}, pp. \bibinfo{pages}{28--39}.
\bibitem[{Losada et~al.(2017)Losada, Crestani, and Parapar}]{losada2017clef}
\bibinfo{author}{D.~E. Losada}, \bibinfo{author}{F.~Crestani}, \bibinfo{author}{J.~Parapar},
\newblock \bibinfo{title}{Clef 2017 erisk overview: Early risk prediction on the internet: Experimental foundations.},
\newblock \bibinfo{journal}{CLEF (Working Notes)} \bibinfo{volume}{850} (\bibinfo{year}{2017}).
\bibitem[{Busch(2009)}]{busch2009anger}
\bibinfo{author}{F.~N. Busch},
\newblock \bibinfo{title}{Anger and depression},
\newblock \bibinfo{journal}{Advances in psychiatric treatment} \bibinfo{volume}{15} (\bibinfo{year}{2009}) \bibinfo{pages}{271--278}.
\bibitem[{Jarema et~al.(2014)Jarema, Dudek, and Czernikiewicz}]{jarema2014cognitive}
\bibinfo{author}{M.~Jarema}, \bibinfo{author}{D.~Dudek}, \bibinfo{author}{A.~Czernikiewicz},
\newblock \bibinfo{title}{Cognitive dysfunctions in depression--underestimated symptom or new dimension},
\newblock \bibinfo{journal}{Psychiatr Pol} \bibinfo{volume}{48} (\bibinfo{year}{2014}) \bibinfo{pages}{1105--16}.
\bibitem[{D’Agostino et~al.(2020)D’Agostino, Pepi, Monti, and Starcevic}]{d2020feeling}
\bibinfo{author}{A.~D’Agostino}, \bibinfo{author}{R.~Pepi}, \bibinfo{author}{M.~R. Monti}, \bibinfo{author}{V.~Starcevic},
\newblock \bibinfo{title}{The feeling of emptiness: a review of a complex subjective experience},
\newblock \bibinfo{journal}{Harvard review of psychiatry} \bibinfo{volume}{28} (\bibinfo{year}{2020}) \bibinfo{pages}{287--295}.
\bibitem[{Abramson et~al.(1989)Abramson, Metalsky, and Alloy}]{abramson1989hopelessness}
\bibinfo{author}{L.~Y. Abramson}, \bibinfo{author}{G.~I. Metalsky}, \bibinfo{author}{L.~B. Alloy},
\newblock \bibinfo{title}{Hopelessness depression: A theory-based subtype of depression.},
\newblock \bibinfo{journal}{Psychological review} \bibinfo{volume}{96} (\bibinfo{year}{1989}) \bibinfo{pages}{358}.
\bibitem[{Erzen and {\c{C}}ikrikci(2018)}]{erzen2018effect}
\bibinfo{author}{E.~Erzen}, \bibinfo{author}{{\"O}.~{\c{C}}ikrikci},
\newblock \bibinfo{title}{The effect of loneliness on depression: A meta-analysis},
\newblock \bibinfo{journal}{International Journal of Social Psychiatry} \bibinfo{volume}{64} (\bibinfo{year}{2018}) \bibinfo{pages}{427--435}.
\bibitem[{Mouchet-Mages and Bayl{\'e}(2008)}]{mouchet2008sadness}
\bibinfo{author}{S.~Mouchet-Mages}, \bibinfo{author}{F.~J. Bayl{\'e}},
\newblock \bibinfo{title}{Sadness as an integral part of depression},
\newblock \bibinfo{journal}{Dialogues in clinical neuroscience} \bibinfo{volume}{10} (\bibinfo{year}{2008}) \bibinfo{pages}{321--327}.
\bibitem[{Van~Gastel et~al.(1997)Van~Gastel, Schotte, and Maes}]{vangastel1997prediction}
\bibinfo{author}{A.~Van~Gastel}, \bibinfo{author}{C.~Schotte}, \bibinfo{author}{M.~Maes},
\newblock \bibinfo{title}{The prediction of suicidal intent in depressed patients},
\newblock \bibinfo{journal}{Acta Psychiatrica Scandinavica} \bibinfo{volume}{96} (\bibinfo{year}{1997}) \bibinfo{pages}{254--259}.
\bibitem[{Zahn et~al.(2015)Zahn, Lythe, Gethin, Green, Deakin, Young, and Moll}]{zahn2015selfblame}
\bibinfo{author}{R.~Zahn}, \bibinfo{author}{K.~E. Lythe}, \bibinfo{author}{J.~A. Gethin}, \bibinfo{author}{S.~Green}, \bibinfo{author}{J.~F.~W. Deakin}, \bibinfo{author}{A.~H. Young}, \bibinfo{author}{J.~Moll},
\newblock \bibinfo{title}{The role of self-blame and worthlessness in the psychopathology of major depressive disorder},
\newblock \bibinfo{journal}{Journal of Affective Disorders} \bibinfo{volume}{186} (\bibinfo{year}{2015}) \bibinfo{pages}{337--341}.
\bibitem[{Ta et~al.(2023)Ta, Rahman, Majumder, Hussain, Najjar, Howard, Poria, and Gelbukh}]{ta2023wikides}
\bibinfo{author}{H.~T. Ta}, \bibinfo{author}{A.~B.~S. Rahman}, \bibinfo{author}{N.~Majumder}, \bibinfo{author}{A.~Hussain}, \bibinfo{author}{L.~Najjar}, \bibinfo{author}{N.~Howard}, \bibinfo{author}{S.~Poria}, \bibinfo{author}{A.~Gelbukh},
\newblock \bibinfo{title}{Wikides: A wikipedia-based dataset for generating short descriptions from paragraphs},
\newblock \bibinfo{journal}{Information Fusion} \bibinfo{volume}{90} (\bibinfo{year}{2023}) \bibinfo{pages}{265--282}.
\bibitem[{Bennett et~al.(1954)Bennett, Alpert, and Goldstein}]{bennett1954communications}
\bibinfo{author}{E.~M. Bennett}, \bibinfo{author}{R.~Alpert}, \bibinfo{author}{A.~Goldstein},
\newblock \bibinfo{title}{{Communications through limited-response questioning}},
\newblock \bibinfo{journal}{Public Opinion Quarterly} \bibinfo{volume}{18} (\bibinfo{year}{1954}) \bibinfo{pages}{303--308}.
\bibitem[{Davies and Fleiss(1982)}]{davies1982measuring}
\bibinfo{author}{M.~Davies}, \bibinfo{author}{J.~L. Fleiss},
\newblock \bibinfo{title}{{Measuring agreement for multinomial data}},
\newblock \bibinfo{journal}{Biometrics}  (\bibinfo{year}{1982}) \bibinfo{pages}{1047--1051}.
\bibitem[{Klaus(1980)}]{klaus1980content}
\bibinfo{author}{K.~Klaus}, \bibinfo{title}{{Content analysis: An introduction to its methodology}}, \bibinfo{year}{1980}.
\bibitem[{Scott(1955)}]{scott1955reliability}
\bibinfo{author}{W.~A. Scott},
\newblock \bibinfo{title}{{Reliability of content analysis: The case of nominal scale coding}},
\newblock \bibinfo{journal}{Public opinion quarterly}  (\bibinfo{year}{1955}) \bibinfo{pages}{321--325}.
\bibitem[{Zapf et~al.(2016)Zapf, Castell, Morawietz, and Karch}]{zapf2016measuring}
\bibinfo{author}{A.~Zapf}, \bibinfo{author}{S.~Castell}, \bibinfo{author}{L.~Morawietz}, \bibinfo{author}{A.~Karch},
\newblock \bibinfo{title}{{Measuring inter-rater reliability for nominal data--which coefficients and confidence intervals are appropriate?}},
\newblock \bibinfo{journal}{BMC medical research methodology} \bibinfo{volume}{16} (\bibinfo{year}{2016}) \bibinfo{pages}{1--10}.
\bibitem[{Craig(1981)}]{craig1981generalization}
\bibinfo{author}{R.~T. Craig},
\newblock \bibinfo{title}{{Generalization of Scott's index of intercoder agreement}},
\newblock \bibinfo{journal}{Public Opinion Quarterly} \bibinfo{volume}{45} (\bibinfo{year}{1981}) \bibinfo{pages}{260--264}.
\bibitem[{Warrens(2012)}]{warrens2012effect}
\bibinfo{author}{M.~J. Warrens},
\newblock \bibinfo{title}{{The effect of combining categories on Bennett, Alpert and Goldstein’s S}},
\newblock \bibinfo{journal}{Statistical Methodology} \bibinfo{volume}{9} (\bibinfo{year}{2012}) \bibinfo{pages}{341--352}.
\bibitem[{McHugh(2012)}]{mchugh2012interrater}
\bibinfo{author}{M.~L. McHugh},
\newblock \bibinfo{title}{{Interrater reliability: the kappa statistic}},
\newblock \bibinfo{journal}{Biochemia medica} \bibinfo{volume}{22} (\bibinfo{year}{2012}) \bibinfo{pages}{276--282}.
\bibitem[{Boyd et~al.(2022)Boyd, Ashokkumar, Seraj, and Pennebaker}]{boyd2022development}
\bibinfo{author}{R.~L. Boyd}, \bibinfo{author}{A.~Ashokkumar}, \bibinfo{author}{S.~Seraj}, \bibinfo{author}{J.~W. Pennebaker},
\newblock \bibinfo{title}{The development and psychometric properties of liwc-22},
\newblock \bibinfo{journal}{Austin, TX: University of Texas at Austin}  (\bibinfo{year}{2022}) \bibinfo{pages}{1--47}.
\bibitem[{Nassar et~al.(2022)Nassar, Helmy, and Ramadan}]{nassar2022binary}
\bibinfo{author}{R.~Nassar}, \bibinfo{author}{A.~Helmy}, \bibinfo{author}{N.~Ramadan},
\newblock \bibinfo{title}{Binary labeled depression corpus of 60,000 english tweets},
\newblock \bibinfo{journal}{Harvard Dataverse}  (\bibinfo{year}{2022}).
\bibitem[{Demszky et~al.(2020)Demszky, Movshovitz-Attias, Ko, Cowen, Nemade, and Ravi}]{demszky2020goemotions}
\bibinfo{author}{D.~Demszky}, \bibinfo{author}{D.~Movshovitz-Attias}, \bibinfo{author}{J.~Ko}, \bibinfo{author}{A.~Cowen}, \bibinfo{author}{G.~Nemade}, \bibinfo{author}{S.~Ravi},
\newblock \bibinfo{title}{Goemotions: A dataset of fine-grained emotions},
\newblock \bibinfo{journal}{arXiv preprint arXiv:2005.00547}  (\bibinfo{year}{2020}).
\bibitem[{Nassar et~al.(2022)Nassar, Helmy, and Ramadan}]{radwa2022multi}
\bibinfo{author}{R.~Nassar}, \bibinfo{author}{A.~Helmy}, \bibinfo{author}{N.~Ramadan}, \bibinfo{title}{Multi labeled depression corpus of 57,000 english tweets}, \bibinfo{year}{2022}. \URLprefix \url{https://doi.org/10.7910/DVN/CMDF4U}. \DOIprefix\doi{10.7910/DVN/CMDF4U}.
\bibitem[{Haque et~al.(2021)Haque, Reddi, and Giallanza}]{haque2021deep}
\bibinfo{author}{A.~Haque}, \bibinfo{author}{V.~Reddi}, \bibinfo{author}{T.~Giallanza},
\newblock \bibinfo{title}{Deep learning for suicide and depression identification with unsupervised label correction},
\newblock in: \bibinfo{booktitle}{Artificial Neural Networks and Machine Learning--ICANN 2021: 30th International Conference on Artificial Neural Networks, Bratislava, Slovakia, September 14--17, 2021, Proceedings, Part V 30}, \bibinfo{organization}{Springer}, \bibinfo{year}{2021}, pp. \bibinfo{pages}{436--447}.
\bibitem[{Joachims(1998)}]{joachims1998text}
\bibinfo{author}{T.~Joachims},
\newblock \bibinfo{title}{Text categorization with support vector machines: Learning with many relevant features},
\newblock in: \bibinfo{booktitle}{European conference on machine learning}, \bibinfo{organization}{Springer}, \bibinfo{year}{1998}, pp. \bibinfo{pages}{137--142}.
\bibitem[{Chen and Guestrin(2016)}]{chen2016xgboost}
\bibinfo{author}{T.~Chen}, \bibinfo{author}{C.~Guestrin},
\newblock \bibinfo{title}{Xgboost: A scalable tree boosting system},
\newblock in: \bibinfo{booktitle}{Proceedings of the 22nd acm sigkdd international conference on knowledge discovery and data mining}, \bibinfo{year}{2016}, pp. \bibinfo{pages}{785--794}.
\bibitem[{Patle and Chouhan(2013)}]{patle2013svm}
\bibinfo{author}{A.~Patle}, \bibinfo{author}{D.~S. Chouhan},
\newblock \bibinfo{title}{Svm kernel functions for classification},
\newblock in: \bibinfo{booktitle}{2013 International conference on advances in technology and engineering (ICATE)}, \bibinfo{organization}{IEEE}, \bibinfo{year}{2013}, pp. \bibinfo{pages}{1--9}.
\bibitem[{Ta et~al.(2022)Ta, Rahman, Najjar, and Gelbukh}]{ta2022gan}
\bibinfo{author}{H.~T. Ta}, \bibinfo{author}{A.~B.~S. Rahman}, \bibinfo{author}{L.~Najjar}, \bibinfo{author}{A.~Gelbukh},
\newblock \bibinfo{title}{Gan-bert: Adversarial learning for detection of aggressive and violent incidents from social media},
\newblock in: \bibinfo{booktitle}{Proceedings of the Iberian Languages Evaluation Forum (IberLEF 2022), CEUR Workshop Proceedings. CEUR-WS. org}, \bibinfo{year}{2022}.
\bibitem[{Zhu et~al.(2023)Zhu, Zhang, Haq, Hui, and Tyson}]{zhu2023can}
\bibinfo{author}{Y.~Zhu}, \bibinfo{author}{P.~Zhang}, \bibinfo{author}{E.-U. Haq}, \bibinfo{author}{P.~Hui}, \bibinfo{author}{G.~Tyson},
\newblock \bibinfo{title}{Can chatgpt reproduce human-generated labels? a study of social computing tasks},
\newblock \bibinfo{journal}{arXiv preprint arXiv:2304.10145}  (\bibinfo{year}{2023}).
\bibitem[{Gilardi et~al.(2023)Gilardi, Alizadeh, and Kubli}]{gilardi2023chatgpt}
\bibinfo{author}{F.~Gilardi}, \bibinfo{author}{M.~Alizadeh}, \bibinfo{author}{M.~Kubli},
\newblock \bibinfo{title}{Chatgpt outperforms crowd-workers for text-annotation tasks},
\newblock \bibinfo{journal}{arXiv preprint arXiv:2303.15056}  (\bibinfo{year}{2023}).
\bibitem[{Belal et~al.(2023)Belal, She, and Wong}]{belal2023leveraging}
\bibinfo{author}{M.~Belal}, \bibinfo{author}{J.~She}, \bibinfo{author}{S.~Wong},
\newblock \bibinfo{title}{Leveraging chatgpt as text annotation tool for sentiment analysis},
\newblock \bibinfo{journal}{arXiv preprint arXiv:2306.17177}  (\bibinfo{year}{2023}).

\end{thebibliography}


\end{document}